\documentclass{article}

\usepackage{arxiv}

\usepackage[utf8]{inputenc} 
\usepackage[T1]{fontenc}    
\usepackage[hidelinks]{hyperref}       
\usepackage{url}            
\usepackage{booktabs}       
\usepackage{amsfonts}       
\usepackage{nicefrac}       
\usepackage{microtype}      
\usepackage{lipsum}		
\usepackage{graphicx}
\usepackage[square,numbers]{natbib}
\usepackage{doi}
\usepackage{multirow}%
\usepackage{amsmath,amssymb,amsfonts}%
\usepackage{amsthm}%
\usepackage{mat hrsfs}%
\usepackage[title]{appendix}%
\usepackage{textcomp}%
\usepackage{manyfoot}%
\usepackage{listings}%
\usepackage{adjustbox}
\usepackage{colortbl}
\usepackage{enumitem}
\usepackage{wrapfig}
\usepackage[most]{tcolorbox}
\usepackage{fancyvrb,fvextra}
\usepackage{subcaption}
\usepackage{xcolor}
\usepackage{float}
\usepackage{makecell}
\usepackage{soul} 

\usepackage{xspace} 

\usepackage[scaled=0.8]{FiraMono}
\definecolor{codegreen}{rgb}{0,0.6,0}
\definecolor{codegray}{rgb}{0.5,0.5,0.5}
\definecolor{codepurple}{rgb}{0.58,0,0.82}
\lstdefinestyle{codestyle}{
    commentstyle=\color{codegreen},
    keywordstyle=\color{magenta},
    numberstyle=\tiny\color{codegray},
    stringstyle=\color{codepurple},
    basicstyle=\ttfamily\small
}
\definecolor{degradation}{HTML}{F6D8D8}

\sethlcolor{degradation}  

\newcommand{\symbolimg}[2][0.3cm]{%
  \ensuremath{\vcenter{\hbox{\includegraphics[height=#1]{#2}}}}%
}

\newcolumntype{/}{!{\color{white}\vline width 7pt}}

\newcommand{\full}{\symbolimg[0.35cm]{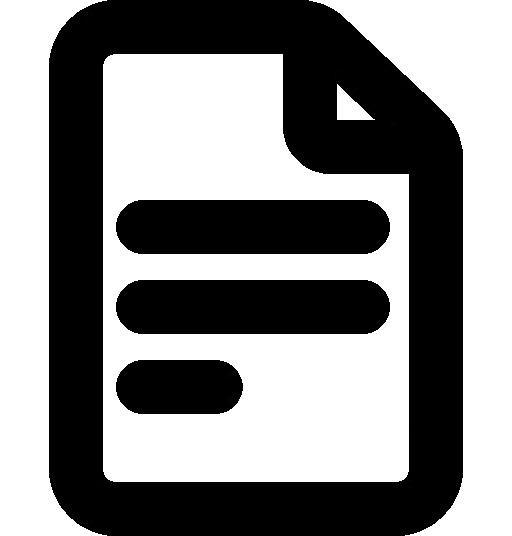}}
\newcommand{\concat}{\symbolimg[0.35cm]{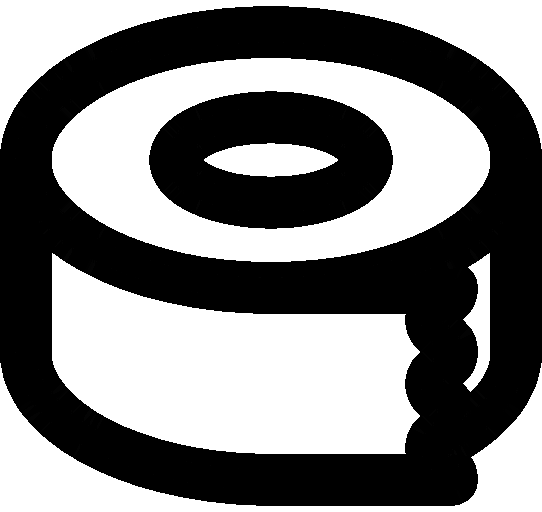}}
\newcommand{\sharded}{\symbolimg[0.35cm]{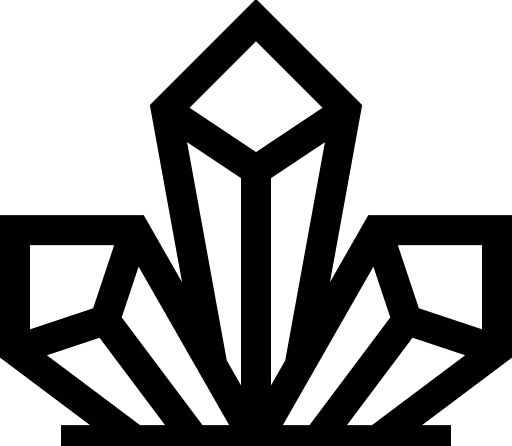}}
\newcommand{\snowball}{\symbolimg[0.35cm]{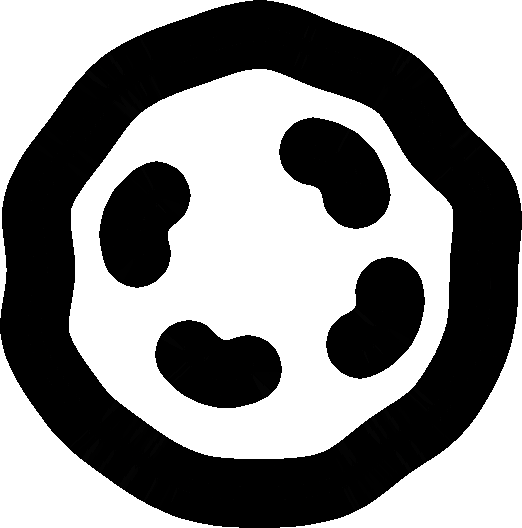}}
\newcommand{\recap}{\symbolimg[0.35cm]{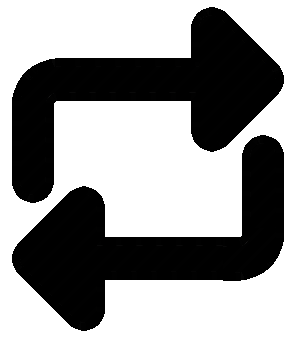}}

\newcommand{\fullt}{\textsc{Full}\xspace}
\newcommand{\concatt}{\textsc{Concat}\xspace}
\newcommand{\shardedt}{\textsc{Sharded}\xspace}
\newcommand{\snowballt}{\textsc{Snowball}\xspace}
\newcommand{\recapt}{\textsc{Recap}\xspace}

\newcommand{\llama}{\symbolimg[0.35cm]{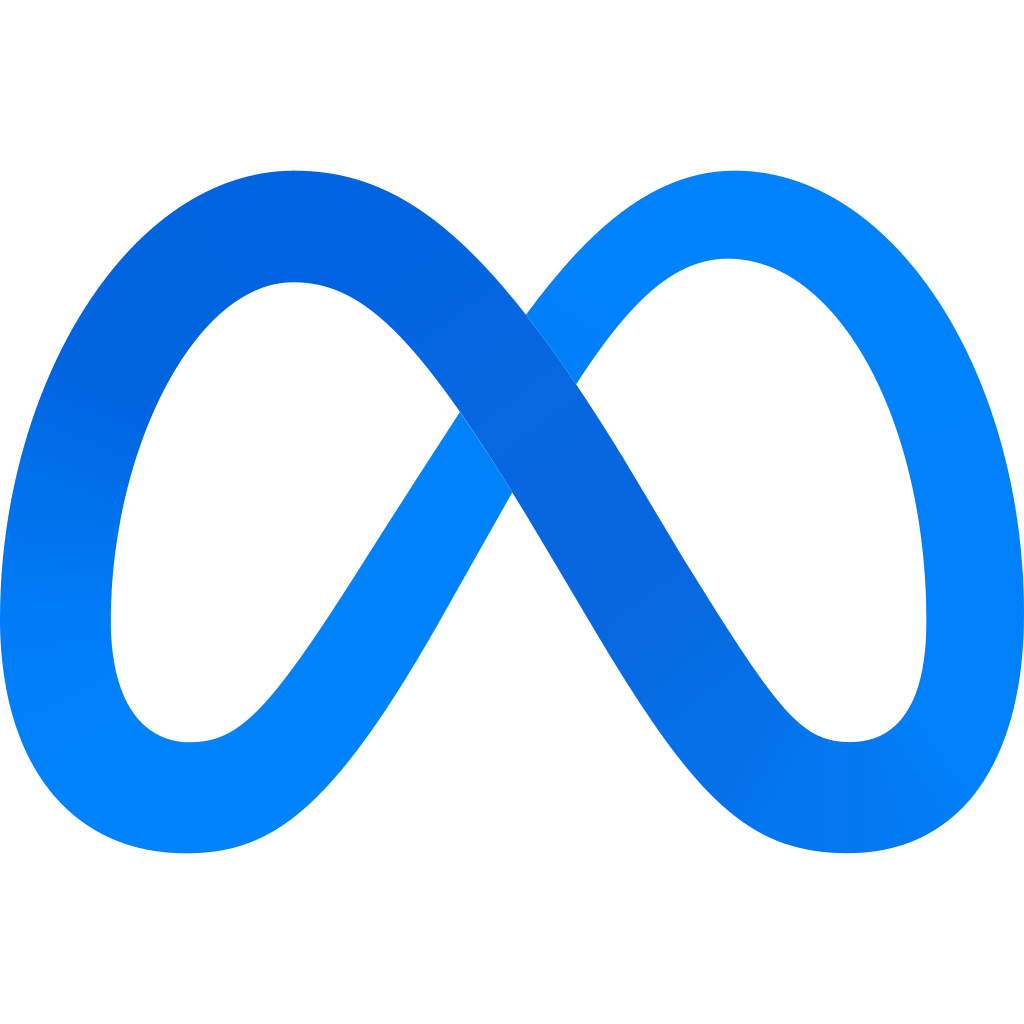}}
\newcommand{\gemini}{\symbolimg[0.35cm]{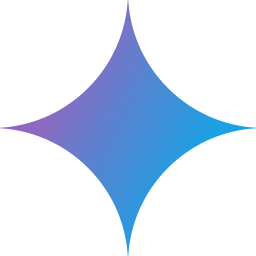}}
\newcommand{\claude}{\symbolimg[0.35cm]{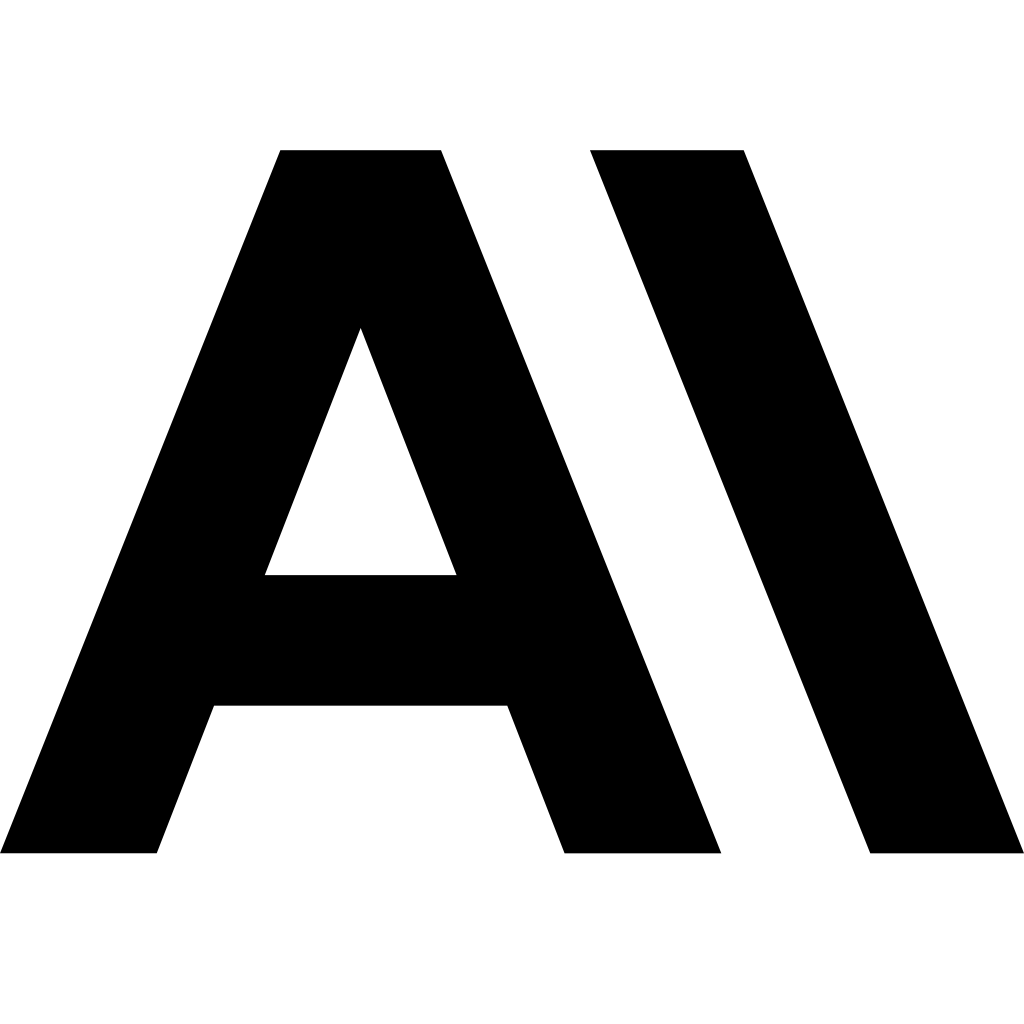}}
\newcommand{\openai}{\symbolimg[0.35cm]{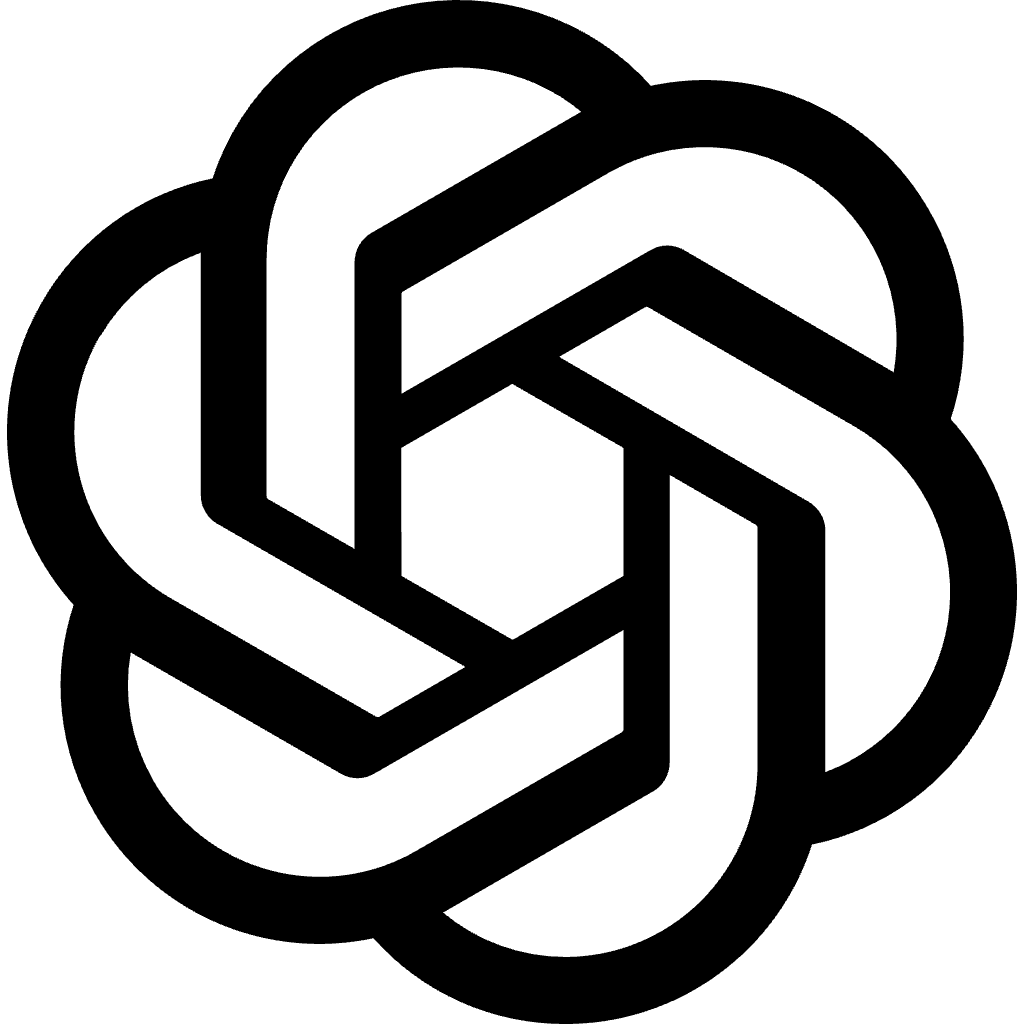}}
\newcommand{\deepseek}{\symbolimg[0.35cm]{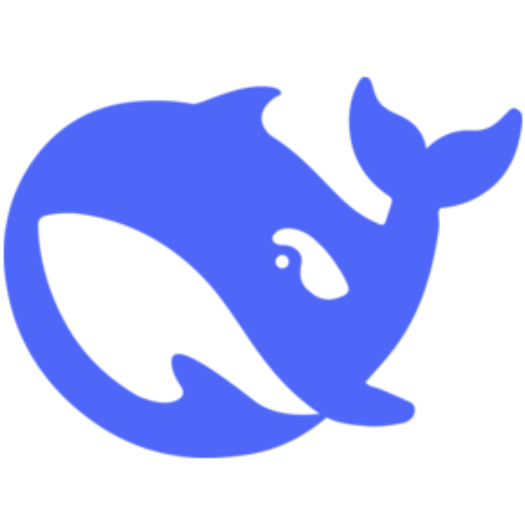}}
\newcommand{\microsoft}{\symbolimg[0.3cm]{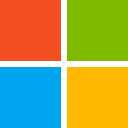}}
\newcommand{\aitwo}{\symbolimg[0.3cm]{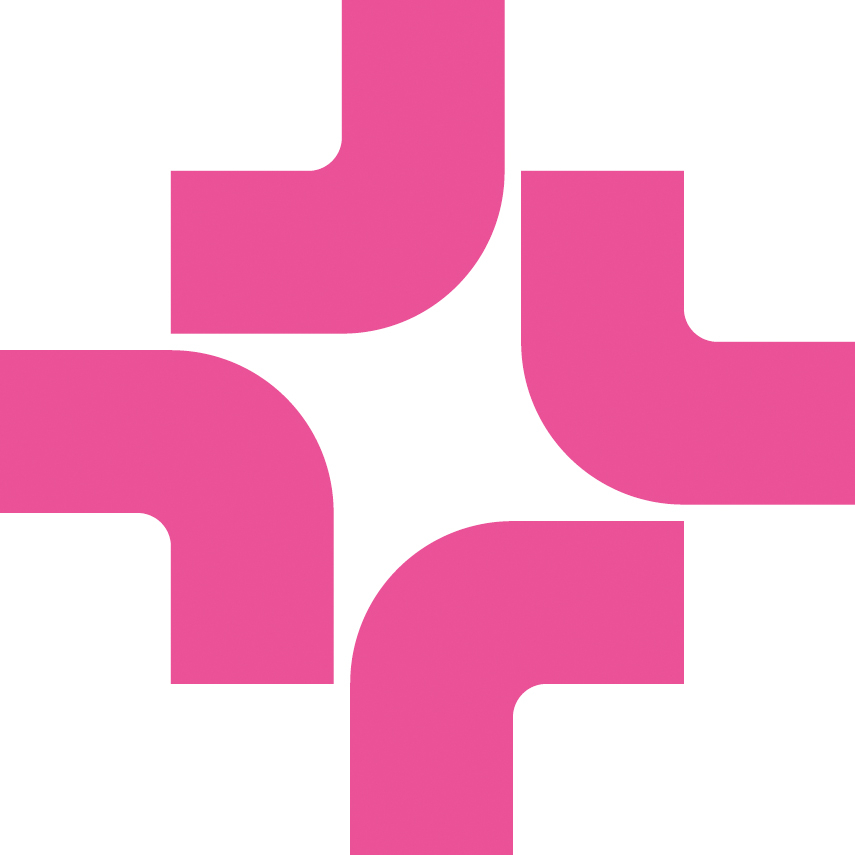}}
\newcommand{\cohere}{\symbolimg[0.3cm]{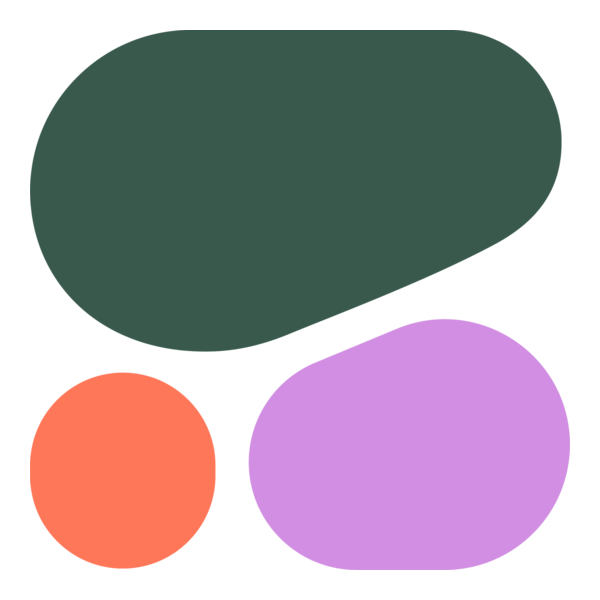}}

\newcommand{\huggingface}{\symbolimg[0.3cm]{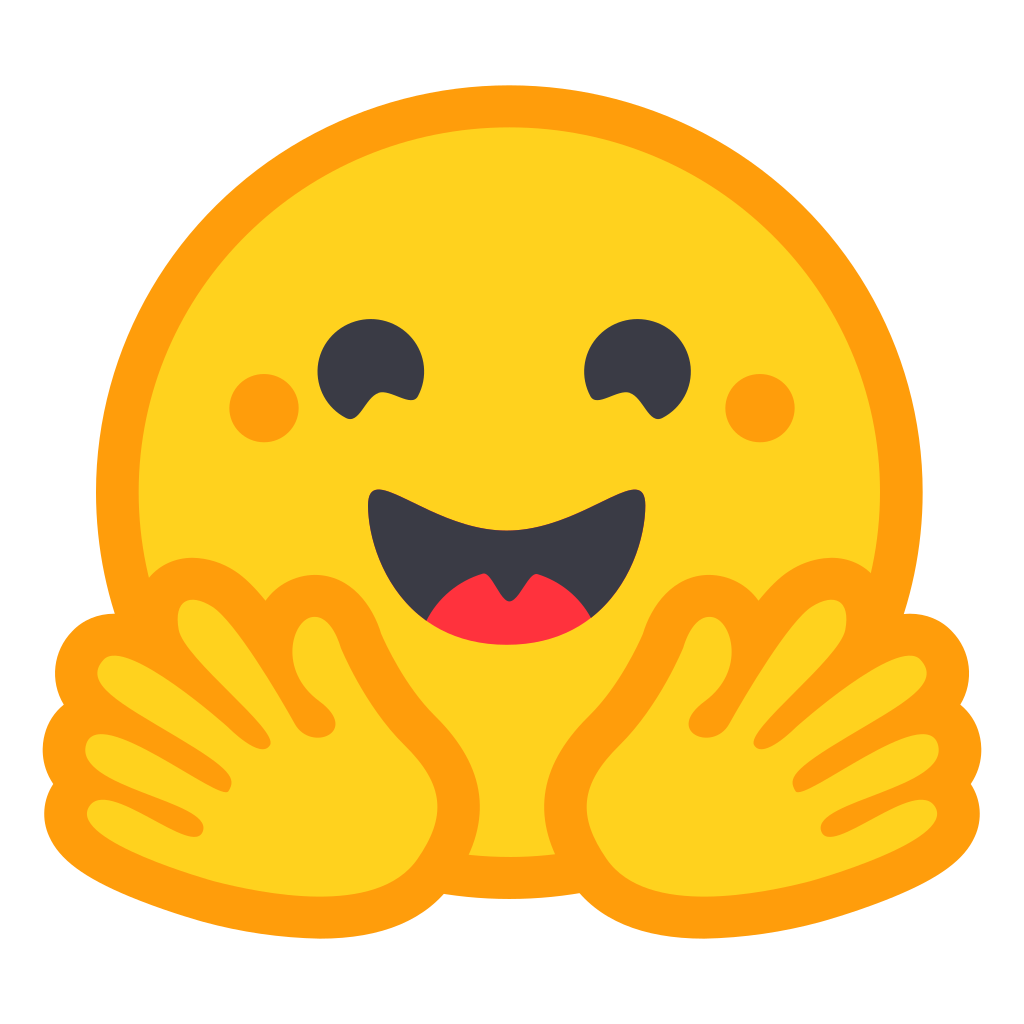}}
\newcommand{\github}{\symbolimg[0.3cm]{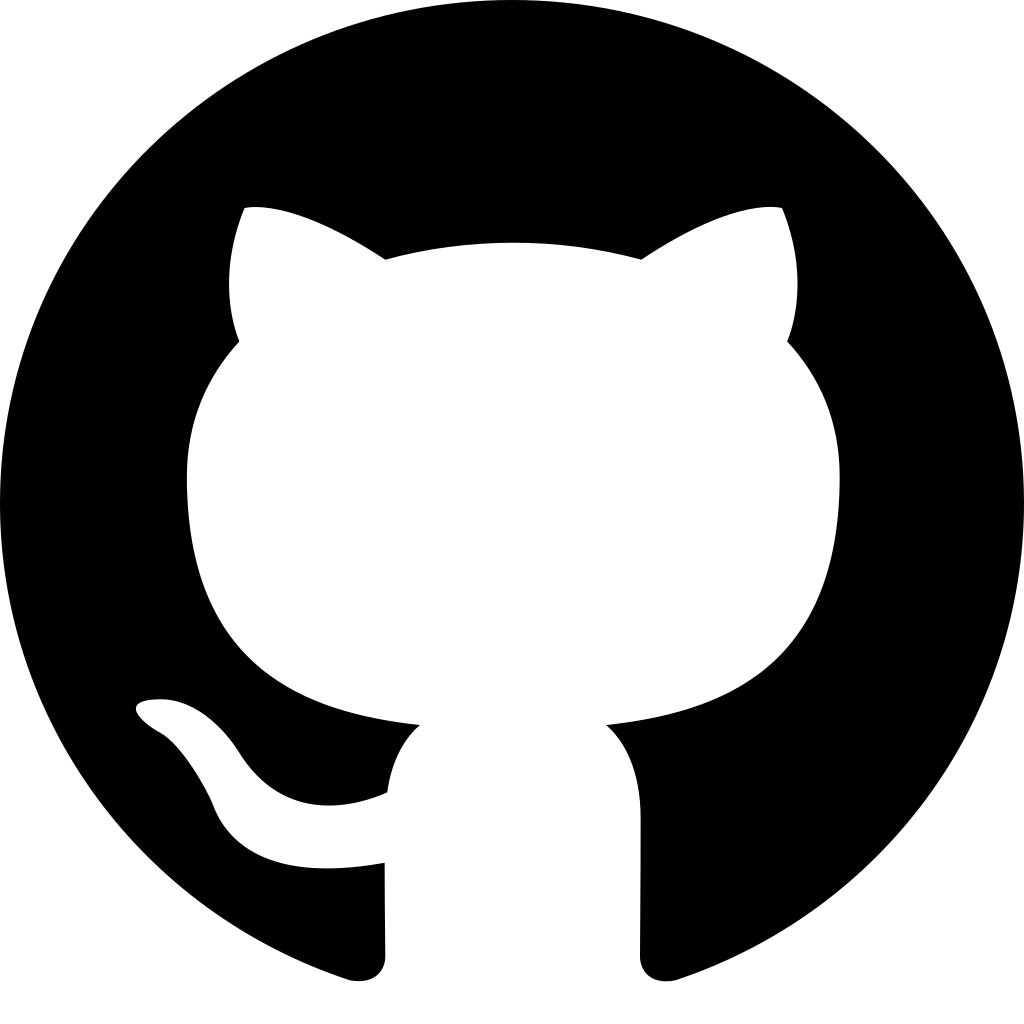}}

\newcommand{\apis}{\symbolimg[0.35cm]{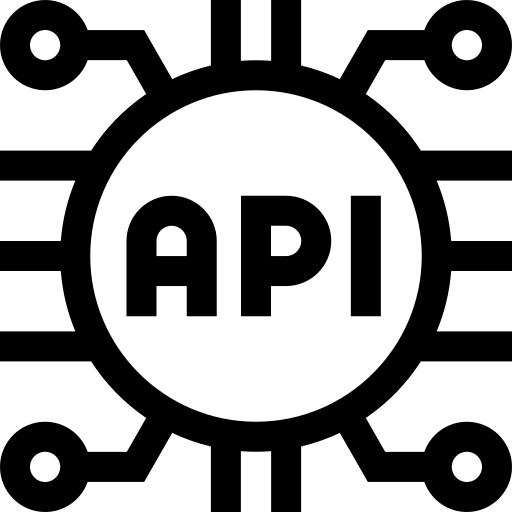}\hspace{0.2em}}
\newcommand{\datattext}{\symbolimg[0.35cm]{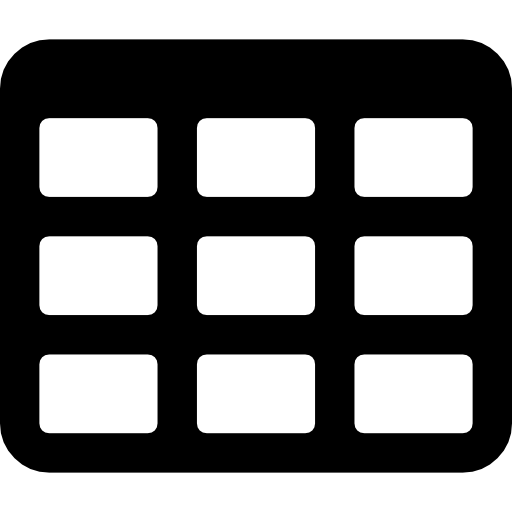}\hspace{0.2em}}
\newcommand{\database}{\symbolimg[0.35cm]{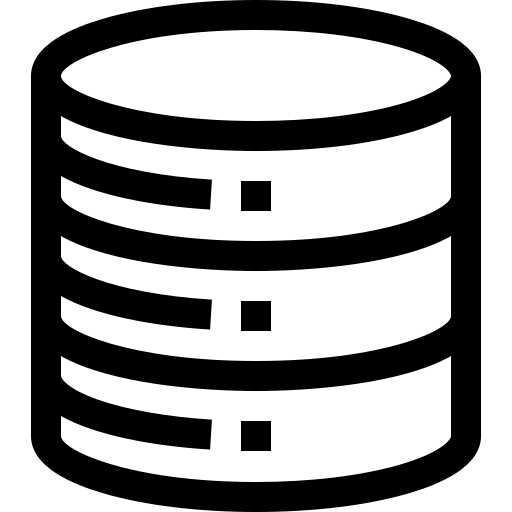}\hspace{0.2em}}
\newcommand{\tmath}{\symbolimg[0.35cm]{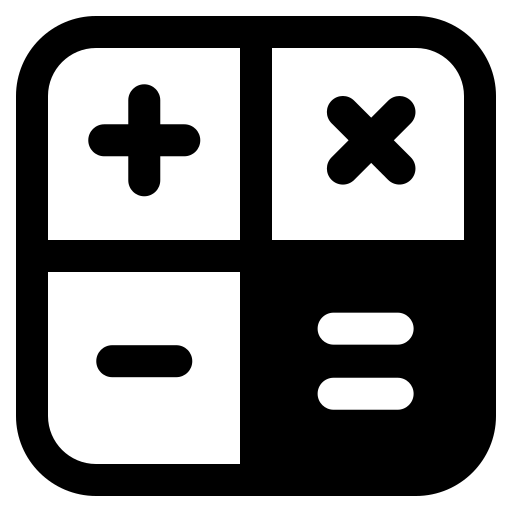}\hspace{0.2em}}
\newcommand{\python}{\symbolimg[0.35cm]{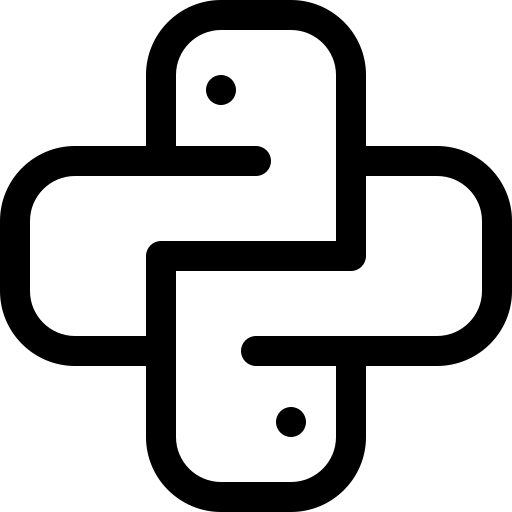}\hspace{0.2em}}
\newcommand{\summary}{\symbolimg[0.35cm]{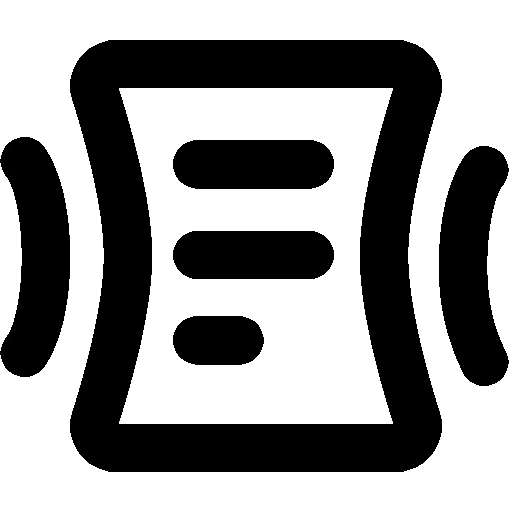}\hspace{0.2em}}
\newcommand{\translation}{\symbolimg[0.35cm]{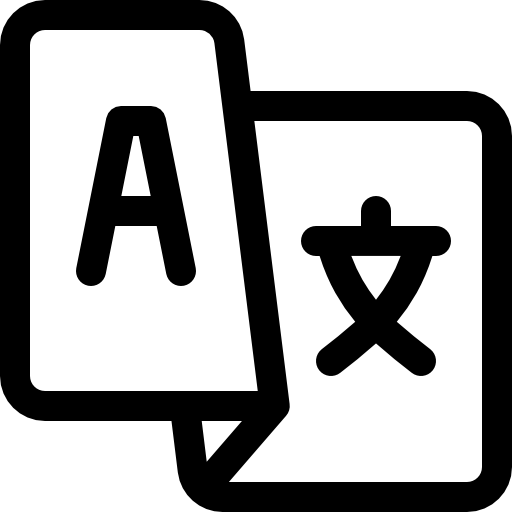}\hspace{0.2em}}

\newcommand{\ccol}[1]{\cellcolor[HTML]{#1}}
\definecolor{coolblue}{HTML}{6D9EEB}
\definecolor{warmgold}{HTML}{F1C232}

\newcommand{\shortp}[1]{\vspace{-1em}\paragraph{#1}}

\newcommand{\jn}[1]{}
\renewcommand{\jn}[1]{{\color{red} JN: {#1}}}

\newcommand{\hh}[1]{}
\renewcommand{\hh}[1]{{\color{blue} HH: {#1}}}

\title{LLMs Get Lost In Multi-Turn Conversation}

\author{
Philippe Laban\thanks{Equal Contributions}~~$^\diamondsuit$ \qquad Hiroaki Hayashi\textsuperscript{$*$}$^\clubsuit$ \qquad Yingbo Zhou$^\clubsuit$ \qquad Jennifer Neville$^\diamondsuit$\\
	$^\diamondsuit$Microsoft Research \qquad $^\clubsuit$Salesforce Research\\
	\texttt{\{plaban,jenneville\}@microsoft.com} \\
	\texttt{\{hiroakihayashi,yingbo.zhou\}@salesforce.com} \\
}

\date{}

\hypersetup{
pdftitle={LLMs Get Lost In Multi-Turn Conversation},
}

\begin{document}
\vspace{-2ex}
\maketitle

\vspace{-5ex}
\begin{abstract}
Large Language Models (LLMs) are conversational interfaces. As such, LLMs have the potential to assist their users not only when they can fully specify the task at hand, but also to help them define, explore, and refine what they need through multi-turn conversational exchange. Although analysis of LLM conversation logs has confirmed that underspecification occurs frequently in user instructions, LLM evaluation has predominantly focused on the single-turn, fully-specified instruction setting. In this work, we perform large-scale simulation experiments to compare LLM performance in single- and multi-turn settings. Our experiments confirm that all the top open- and closed-weight LLMs we test exhibit significantly lower performance in multi-turn conversations than single-turn, with an average drop of 39\% across six generation tasks. Analysis of 200,000+ simulated conversations decomposes the performance degradation into two components: a minor loss in aptitude and a significant increase in unreliability. We find that LLMs often make assumptions in early turns and prematurely attempt to generate final solutions, on which they overly rely. In simpler terms, we discover that \textbf{when LLMs take a wrong turn in a conversation, they get lost and do not recover}.
\end{abstract}

\vspace{-1ex}
\begin{center}
\href{https://github.com/Microsoft/lost_in_conversation}{\github~\texttt{Microsoft/lost\_in\_conversation}} \quad
\href{https://huggingface.co/datasets/Microsoft/lost_in_conversation}{\huggingface~\texttt{datasets/Microsoft/lost\_in\_conversation}}
\end{center}
\vspace{-2ex}

\begin{figure}[ht]
    \centering
    \includegraphics[width=1.0\linewidth]{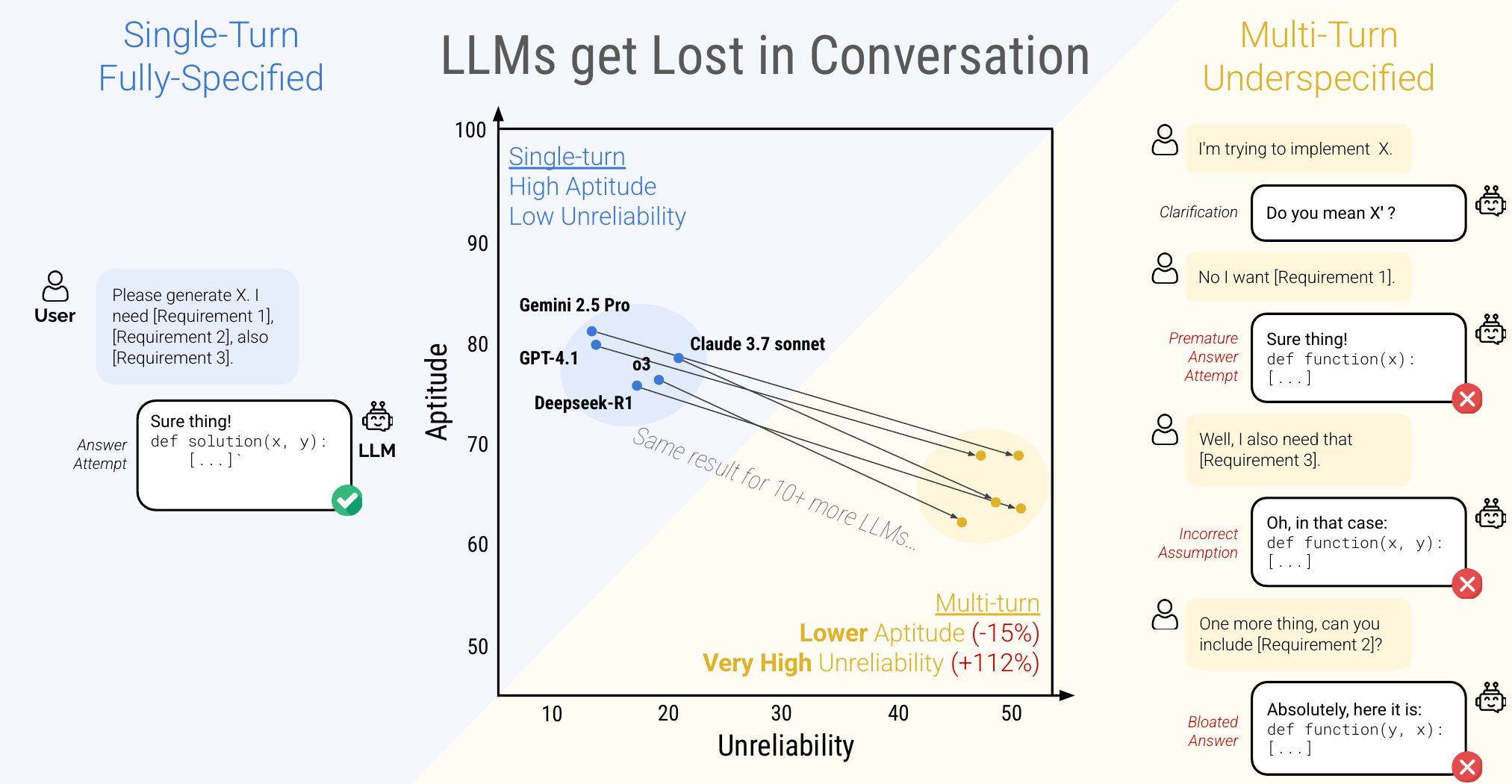}
    \caption{In this work, we simulate single- and multi-turn conversations for six generation tasks. The 15 LLMs we test perform much worse in multi-turn settings (-35\%) explained by some loss in aptitude, and large losses in reliability. Aptitude is defined as performance in best-case conversation simulation, and unreliability as the gap between best- and worst-case performance. In short, we find that LLMs get lost in multi-turn, underspecified conversation.}
    \label{fig:teaser}
    \vspace{-1ex}
\end{figure}


\section{Introduction} \label{sec:intro}

Today's large language models (LLMs) function as conversational interfaces (\textit{e.g.}, ChatGPT, Gemini, Claude), enabling users to interact with the LLM through multiple conversation turns. Such interaction promises to help users not only when they know what they need (i.e., they can fully specify their requirements in an instruction), but also when they don't. In such cases, users might start with an underspecified instruction and further clarify their needs through turn interactions.
Though studies of LLM conversation logs have confirmed that underspecification in user instructions is prevalent \citep{herlihy2024overcoming}, LLM systems are typically evaluated in single-turn, fully-specified settings.

Even though a growing body of work proposes to evaluate LLMs in a \textbf{multi-turn} fashion, we identify in our review (Section~\ref{sec:background}) that most prior work treats the conversation as \textit{episodic}: conversation turns might relate to each other, but the conversation can effectively be decomposed as an array of subtasks that can be evaluated in isolation.
We argue that episodic tasks move away from what is prevalent in human conversation: underspecification~\cite{zipf1949human,herlihy2024overcoming}.

In this work, we close this gap by creating a simulation environment for multi-turn underspecified conversations  -- sharded simulation -- that leverages existing instructions from high-quality single-turn benchmarks. At a high level, the sharding process we propose transforms existing single-turn instructions into \textit{sharded instructions}, a set of smaller instructions that jointly deliver the same information as the original instruction. Sharded simulation then ensures that each turn of conversation reveals at most one shard of information per conversation turn, enforcing that the instruction is gradually revealed through the conversation.

On the set of tasks that we experimented on, we observed that models engaged in multi-turn underspecified conversations achieved an average performance of 65\%--a 25-point drop from single-turn performances of 90\% when they receive the entire instruction at the beginning of the conversation.
Notably, we observe this drop in performance even in two-turn conversations, and across all LLMs we test, from small open-weights (LLama3.1-8B-Instruct) to state-of-the-art (Gemini 2.5 Pro).

Furthermore, we decompose the performance degradation into two components: (1) loss in aptitude, and (2) increase in unreliability.
We find that in single-turn settings, models with higher aptitude tend to be more reliable (\textit{e.g.}, GPT-4.1, Gemini 2.5 Pro). 
On the other hand, all LLMs exhibit very high unreliability in multi-turn settings, regardless of aptitude.
We refer to this as the \textit{lost in conversation phenomenon}: when LLMs take a wrong turn in multi-turn conversation, they get lost and do not recover.

We investigate several explanations for this effect and show that the LLMs tend to (1) generate overly verbose responses, leading them to (2) propose final solutions prematurely in conversation, (3) make incorrect assumptions about underspecified details, and (4) rely too heavily on previous (incorrect) answer attempts.

Our findings highlight a gap between how LLMs are used in practice and how the models are being evaluated. Ubiquitous performance degradation over multi-turn interactions is likely a reason for low uptake of AI systems \cite{southworth2023developing,Brauner2023WhatDT,horowitz2024adopting}, particularly with novice users who are less skilled at providing complete, detailed instructions from the onset of conversation \cite{zamfirescu2023johnny,Knoth2024AILA}.

The rest of the paper is structured as follows: Section~\ref{sec:background} situates our work with respect to prior work on multi-turn evaluation. In Section~\ref{sec:simulation}, we describe the simulation environment we built for both single- and multi-turn conversations on a diverse set of generation tasks.
We introduce the six tasks and the metrics we use to evaluate the aptitude and reliability of models in Section~\ref{sec:tasks}. Sections~\ref{sec:scale_and_parameters}-\ref{sec:results} define our main experiment involving 15 LLMs, and analyze the main findings.
Finally, the Implications section (Section~\ref{sec:implications}) discusses the ramifications of the work, from the perspective of organizations that are building LLM-based conversation products, to that of end-users of the LLM-based systems.
We provide actionable recommendations based on small-scale experiments and make a concrete call-to-action to LLM builders, urging them to prioritize multi-turn reliability in conjunction with aptitude in future model iterations.

\section{Background and Related Work} \label{sec:background}


Previous-generation language models (e.g., BART \cite{lewis2019bart}, GPT-2 \cite{radford2019language}, or T5 \cite{raffel2020exploring}) were not equipped to handle multi-turn conversations, which led evaluation to focus on single-turn tasks \cite{wang2018glue}. Conversational AI was typically implemented as specialized systems that leveraged language models as components \cite{konrad2021alquist}, and were evaluated through human protocols \cite{deriu2021survey,lee2022evaluating,finch2023don,murakhovs2023salespeople}, or competitions like Amazon's Alex Prize \cite{ram2018conversational}.

As the meteoric rise of ChatGPT led to increased interest in multi-turn evaluation, initial popular efforts such as MT-bench~\cite{zheng2023judging} leveraged crowd-sourced annotations to evaluate LLM-as-a-judge ability.
Follow-up works expanded on MT-bench, for instance to include longer conversations~\cite{kwan2024mt,duan2024botchat}, increase evaluation granularity~\cite{bai2024mt}, or to tackle different aspects such as naturalness~\cite{sirdeshmukh2025multichallenge} or tool use~\cite{2024bfcl,wang2024mint}.

Crucially, such works typically simulate \textit{episodic} conversations: each turn in the conversation introduces a subtask that relates to previous conversation turns, but can be evaluated in isolation.
In this work, we find that episodic tasks overestimate LLM performance in multi-turn conversations (see Section~\ref{sec:implications_nlp}).
In short, although episodic tasks require some level of multi-turn context understanding, they do not involve actively fusing the information to answer \textit{underspecified} user instructions.
Underspecified user instructions are not only common in real-world human-AI communication~\cite{herlihy2024overcoming}, but also a natural tendency in conversations, termed ``the principle of least effort''~\cite{zipf1949human}.
We show that underspecification in multi-turn conversations leads to large and universal performance degradations: LLMs make early assumptions to fill in for missing information, prematurely attempt to propose finalized solutions, and have difficulty adapting and course-correcting when provided with new information.
We make underspecification the central element of our evaluation setting.


Multi-turn episodic evaluation is sometimes framed as a way to evaluate multi-turn model capabilities with higher granularity.
Categories of subtasks (such as refinement, follow-up, expansion, etc.) allow for the study of more specific LLM behavior \cite{bai2024mt, kwan2024mt,sun2024parrot,fan2024fairmt,deng2024multi,liang2024mathchat,han2025can}.
According to such framing, multi-turn tasks \textit{differ} from single-turn tasks and are not evaluated on the same set of tasks.
We argue that this framing is artificial and limits the scope of multi-turn evaluation, restricting the direct comparison of multi-turn and single-turn abilities of LLMs.
In our work, we conduct both single-turn and multi-turn conversation simulations on \textit{a common set of tasks}: controlled experiments that precisely allow us to identify performance degradations from single- to multi-turn settings.

Evaluating LLMs in multi-turn settings is a challenge because conversational trajectories diverge far more than in a single‑turn.
Thus, most previous studies have focused on classification or short-form tasks, with more straightforward evaluation settings.
However, the predominant use cases for LLMs are generative in nature, both for programming (\textit{e.g.}, coding assistants) and natural language (\textit{e.g.}, writing, summarizing) \cite{zheng2023lmsys,handa2025economic}.
Long-form evaluation in the multi-turn setting is therefore essential, as it assesses models' ability to flexibly adapt and refine the response as the users provide more information.
In this work, we focus exclusively on generation tasks that capture widely used scenarios in both programming and natural language domains.


Scaling multi-turn experimentation requires simulating a user.
Existing studies implemented such user simulation in different ways: relying on templates \cite{choi2018quac,reddy2019coqa,laban2023you,deng2024multi}, using an LLM \cite{poelitz2025synthetic,li2024iqa,chang2025chatbench,liang2024mathchat}, involving human annotators \cite{finch2023don,chang2025chatbench}, or real users as part of a study \cite{ram2018conversational,laban2021s,chiang2024chatbot}.
Although involving real users leads to the most natural and realistic conversations, it comes at the cost of scalability and reproducibility.
In this work, we adopt an LLM-based simulator to enable controlled flexibility and divergence.
Nevertheless, a fully automated simulation limits the scope of our findings: the conversations we simulate are not representative of human-AI conversations.
We therefore frame the simulation as a tool to study the \textit{LLM behavior} in the multi-turn setting rather than user behavior.
In addition, as detailed in the Limitations Section (Section~\ref{sec:limitations}), we argue that our simulation framework is simplistic and idealized.
For example, the conversations are guaranteed to end with sufficient information to solve the tasks, and the simulator limits unexpected behavior (\textit{e.g.}, derailing) that can occur in real-world settings.
We suggest these choices imply that degradations observed in this work are most likely underestimates of what occurs in real-world, underspecified multi-turn Human-AI conversations. Appendix~\ref{app:rel_work} introduces other related work specifically focused on underspecified communication.

\section{Simulating Underspecified, Multi-Turn Conversation} \label{sec:simulation}

To assess LLM performance in multi-turn, underspecified conversation, we develop a simulation environment that repurposes existing tasks from single-turn benchmarks. First, we apply a \textit{sharding process} to transform original fully-specified instructions into \textit{sharded instructions}. Second, we implement a sharding simulation environment that carries out a multi-turn conversation based on a sharded instruction.

\subsection{Sharding Process: From Fully-Specified to Sharded Instructions} \label{ssec:sharding_concept}

An original, fully-specified instruction from GSM8K \cite{cobbe2021training} and the equivalent sharded instruction are listed in Figure~\ref{fig:sharding_example}.

The original instruction is a single, long utterance that introduces all the content at once: a high-level question (\textit{i.e.}, ``How long will it take [...]''), context, and conditions. The sharded instruction is composed of \textit{a set of shards}, each introducing a single element from the original instruction. More specifically, the first shard (Shard 1) of a sharded instruction always introduces the high-level intent for the instruction, and subsequent shards each provide clarification to the instruction. Taken jointly, the set of shards reflects the same information provided in the fully-specified instruction, with the information explicitly divided across shards.

In Appendix~\ref{app:sharded_definition}, we provide a more precise and mathematical definition of a sharded instruction in relation to the original fully-specified instruction, and define five key properties a sharded instruction must satisfy to be considered valid.

As part of our work, we developed a semi-automatic sharding process to scale the creation of sharded instructions. This process, described in depth in Appendix~\ref{app:sharding_process}, ensured that the experiments we carried out used sharded instructions that adhered to the properties we defined.

\begin{figure}[t]
  \centering

  \begin{subfigure}[t]{0.38\textwidth}
    \centering
    \begin{tcolorbox}[title=Fully-Specified Instruction (original),colframe=coolblue,colback=coolblue!5,fontupper=\linespread{1.1}\selectfont]
    Jay is making snowballs to prepare for a snowball fight with his sister. He can build 20 snowballs in an hour, but 2 melt every 15 minutes. How long will it take before he has 60 snowballs?
    \end{tcolorbox}
    \subcaption{Original GSM8K instruction.}
    \label{sfig:fully_specified_instruction}
  \end{subfigure}
  \hfill
\begin{subfigure}[t]{0.58\textwidth}
    \centering
    \begin{tcolorbox}[title=Sharded Instruction (based on original), colframe=warmgold, colback=warmgold!5, boxsep=1mm, left=1mm, right=1mm, top=1mm, bottom=1mm]
    Shard 1: How long before Jay’s ready for the snowball fight?

\vspace{-2mm}
\color{warmgold} \rule{\linewidth}{0.4pt}\color{black}

Shard 2: He’s preparing for a snowball fight with his sister.

\vspace{-2mm}
\color{warmgold} \rule{\linewidth}{0.4pt}\color{black}

Shard 3: He can make 20 snowballs per hour.

\vspace{-2mm}
\color{warmgold} \rule{\linewidth}{0.4pt}\color{black}

Shard 4: He’s trying to get to 60 total.

\vspace{-2mm}

\color{warmgold} \rule{\linewidth}{0.4pt}\color{black}

Shard 5: The problem is that 2 melt every 15 minutes.

\vspace{-0.5mm}
\end{tcolorbox}
\subcaption{Equivalent Sharded Instruction.}
\label{sfig:sharded_instruction}
\end{subfigure}

\caption{Paired instructions: (\subref{sfig:fully_specified_instruction}) a fully-specified instruction used in single-turn conversation simulation, and (\subref{sfig:sharded_instruction}) a sharded instruction used to simulate underspecified, multi-turn conversation.}
\label{fig:sharding_example}
\end{figure}

\subsection{Simulating Sharded Conversations} \label{sec:simulation_process}

\begin{figure}[tbhp]
    \centering
    \includegraphics[width=0.9\linewidth]{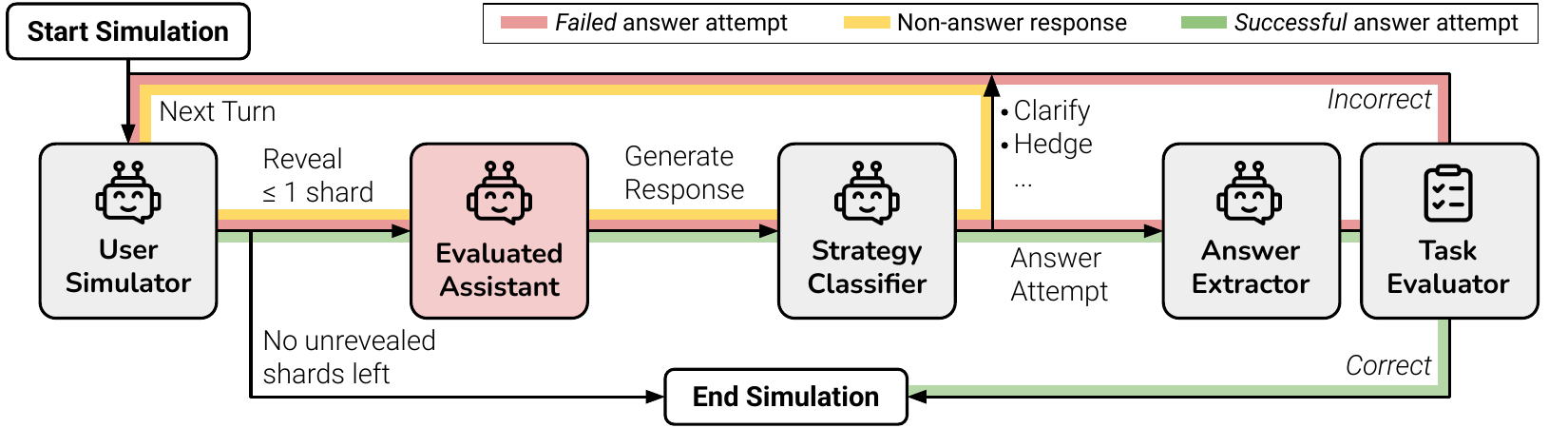}
    \caption{Sharded Conversation Simulation Diagram. The subject for the simulation is highlighted in red.}
    \label{fig:simulator}
\end{figure}

Figure~\ref{fig:simulator} depicts the process of simulating a multi-turn, underspecified conversation based on a sharded instruction. At a high-level, the conversation involves three parties: the \textbf{assistant} is the LLM being evaluated in the simulation, the \textbf{user} (simulated by an LLM) who has access to the entirety of the sharded instruction and is in charge of revealing shards during turns of the conversation, and the \textbf{system} which categorizes and evaluates assistant responses.

On the first turn, the user simulator reveals the first shard of the instruction (\textit{i.e.}, Shard 1) to the assistant, which then generates a free text response. The system processes the assistant's response into one of seven possible response strategies: \textit{clarification}, \textit{refusal}, \textit{hedging}, \textit{interrogation}, \textit{discussion}, \textit{missing}, or \textit{answer attempt},\footnote{See Appendix~\ref{app:answer_categorization} for the definition and the example for each strategy.} based on \citet{herlihy2024overcoming}'s LLM response categorization.
If the assistant generates an answer attempt (\textit{i.e.}, proposing an explicit, full-form solution), then the answer extractor component determines the span that corresponds to the answer within the assistant's free-form response (\textit{e.g.}, code snippet, number). This step is required because LLMs often pad answer attempts with additional information, such as a natural-language explanation or a follow-up question, which could hinder evaluation.
Finally, the extracted answer is scored by a task-specific evaluator function.
Subsequent turns follow a similar pattern: at each turn, the user simulator reveals at most one shard of information, the assistant responds freely, which gets evaluated if the response is classified as an answer attempt. The conversation ends if one of two conditions is met: (1) the task-evaluator assesses that an assistant answer attempt is correct, or (2) if at the start of a new turn, the user simulator has run out of shards to reveal in the conversation.

Preliminary experiments revealed that during simulation, evaluated assistants often asked clarification questions that related to specific shards of the instruction. As such, deciding which shard to reveal next in the conversation (the role of the user simulator) is non-trivial, as it should take into account the state of the conversation so far. We instantiate the user simulator as a low-cost LLM (specifically, GPT-4o-mini) that has access to the entire sharded instruction and the state of the conversation so far, tasking it with deciding the next shard to reveal that fits most naturally in the ongoing simulated conversation. The user simulator is also tasked with rephrasing the shard to fit naturally within the conversation without modifying its informational content. See Appendix~\ref{app:example_conv} for an example simulated sharded conversation.

Besides user messages, the assistant receives a minimal system instruction (before the first turn) that provides the necessary context to accomplish the task (such as a database schema or a list of available API tools). Importantly, the assistant is not explicitly informed that it is participating in a multi-turn, underspecified conversation and is not encouraged to pursue specific conversational strategies. Although such additional instructions would likely alter model behavior, we argue that such changes are not realistic, as such information is not available a priori in practical settings. In summary, we provide no information about the setting to the evaluated assistant model during simulation, aiming to assess default model behavior.

Apart from the user simulator, the strategy classifier and answer extractor components are also implemented with prompt-based GPT-4o-mini. While the choice of LLM-based components in the simulator allows for dynamic choices that provide a more realistic simulation, they also unavoidably lead to simulation errors, which can affect the validity of experiments. To understand the scope of simulation errors and their effect on simulation validity, we conducted an in-depth manual annotation of several hundred simulatesouthworth2023developingd conversations. The annotation effort and its findings are detailed in Appendix~\ref{app:inspection}. In summary, we found that errors introduced by the user simulator, strategy classifier, or answer extraction occurred in less than 5\% of inspected conversations and that these errors disfavored the assistant model in less than 2\% of the conversations. We believe the process described above can accurately simulate multi-turn, underspecified conversations based on sharded instructions, and we rely on it to simulate conversations for our experiments.

\subsection{Simulation Types} \label{sec:five_simulations}

\begin{wrapfigure}{r}{0.5\linewidth}
    \vspace{-6ex}
    \centering
    \includegraphics[width=\linewidth]{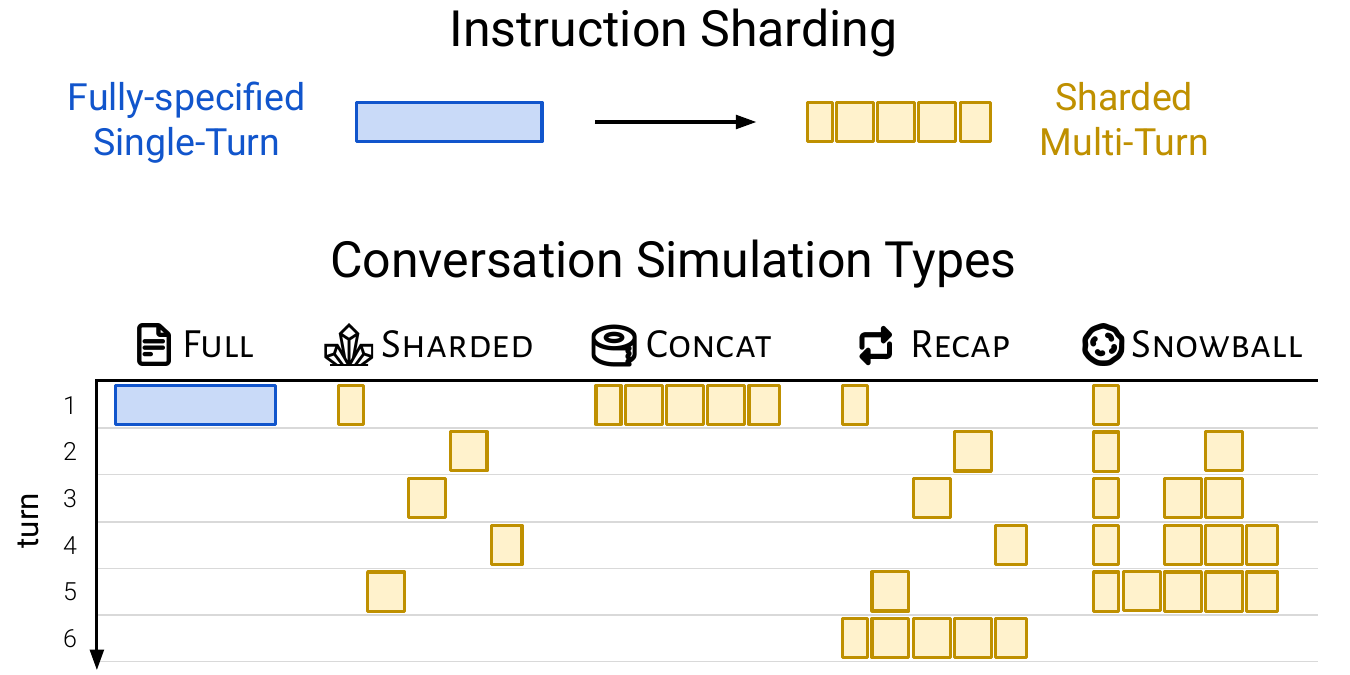}
    \caption{Conversation simulation types based on sharded instructions. Once an original fully-specified instruction (blue block) is sharded (set of yellow blocks), the ``shards'' can be used to simulate single-turn (\textsc{Full}, \textsc{Concat}) or multi-turn (\textsc{Sharded}, \textsc{Recap}, \textsc{Snowball}) conversations, affecting the pace of information disclosure.}
    \label{fig:conversation_types}
    \vspace{-4ex}
\end{wrapfigure}

We leverage sharded instructions to simulate five types of single- or multi-turn conversations, as illustrated in Figure~\ref{fig:conversation_types}. We now introduce each one and explain its purpose in our experiments.

\shortp{\full~\textsc{Fully-specified}} (short-form: \textsc{Full}) simulates single-turn, fully-specified conversations in which the original instruction is provided to the LLM in the first turn. This simulation type evaluates baseline model performance on the tasks.

\shortp{\sharded~\textsc{Sharded}} simulates multi-turn, underspecified conversations as outlined above. \shardedt simulations are our primary tool to evaluate model performance in underspecified, multi-turn conversations.

\shortp{\concat~\textsc{Concat}} simulates single-turn, fully-specified conversation based on the sharded instruction. The shards are concatenated into a single instruction in bullet-point form (with one shard per line), preceded by an instruction to complete the task taking into account all bullet-points. The \concatt simulation is a logical mid-point between full and sharded, in which underspecification is removed (like \fullt) but the rephrasing that occurred during instruction sharding is preserved (like \shardedt). \concatt is intended as a verification baseline: a model that succeeds at both \fullt and \concatt, but not at \shardedt, struggles specifically because of underspecification and the multi-turn nature of the conversation, and not due to the rephrasing that occurred during the sharding process, which may have led to information loss.

\vspace{-1em}
\paragraph{\recap~\textsc{Recap}} simulates a \shardedt conversation, and adds a final \textit{recapitulation turn} which restates all the shards of the instruction in a single turn, giving the LLM one final attempt at responding. \recapt is a combination of the \shardedt simulation followed by a \concatt turn, and is explored as a method in Section~\ref{sec:implications_systems} to evaluate whether such a conceptually simple agent-like intervention can mitigate the loss in performance observed in \shardedt conversations. 

\vspace{-1em}
\paragraph{\snowball~\textsc{Snowball}} takes the \recapt simulation a step further, implementing turn-level recapitulation. At each turn, the user simulator introduces a new shard, but also restates all the shards that have been revealed so far in the conversation, producing a \textit{snowball} effect as each turn reveals all the information from the previous turn, plus one additional shard. The redundancy implemented in the \snowballt simulation is also explored as a method in Section~\ref{sec:implications_systems} to study whether turn-level reminders help alleviate the need for LLMs to recall information across multiple turns of context.

\section{Task and Metric Selection} \label{sec:tasks_and_metrics}

\subsection{Task Selection} \label{sec:tasks}

\begin{figure}[tb]
    \centering
    \includegraphics[width=1.0\linewidth]{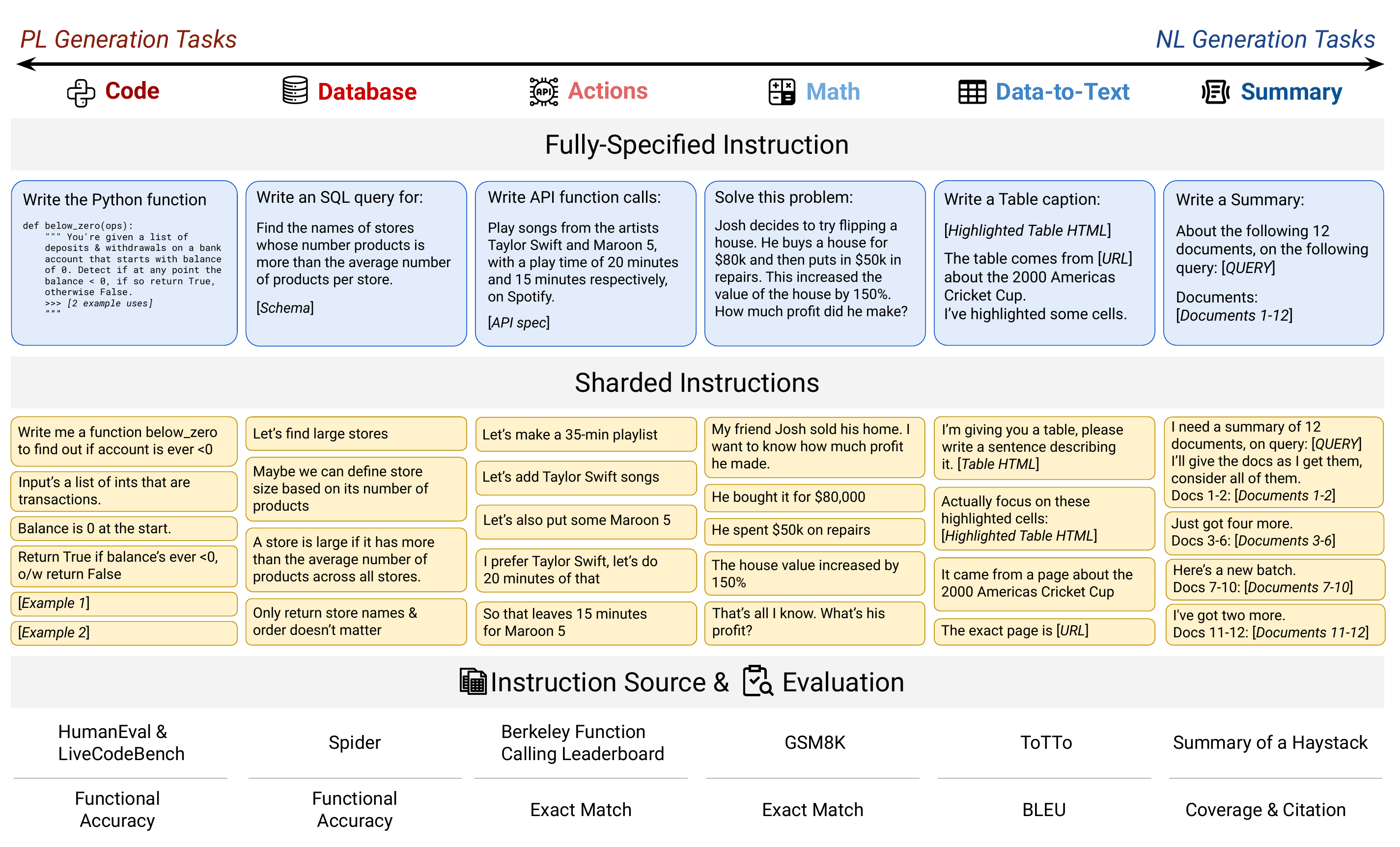}
    \caption{Six sharded tasks included in our experiments. We purposefully include tasks that involve generating programming and natural language. For each task, an illustrative fully-specified instruction and its sharded counterpart. We sharded 90-120 instructions based on high-quality datasets (Instruction Origin), re-purposing existing evaluation.}
    \label{fig:tasks}
    \vspace{-0.2em}
\end{figure}

We constructed sharded instructions for six tasks that we use in a large-scale simulation experiment. For each task, we selected instructions from one or two high-quality single-turn, fully-specified benchmarks, and implemented a semi-automatic sharding process. The process relied first on an LLM (GPT-4o) to propose and verify sharding candidates, which were then reviewed and edited (when necessary) by the authors of the work. The sharding process (outlined in detail in Appendix~\ref{app:sharding_process}) allowed us to scale the construction of sharded instruction corpora while ensuring validity of the underlying instructions. For each task, we prepared 90-120 sharded instructions (each paired with the original single-turn instructions), which required between 1-4 hours of manual inspection and annotation.

We carefully selected popular and diverse generation tasks across programming and non-programming use cases. Figure~\ref{fig:tasks} provides an example of an original and sharded instruction for each task, which we now introduce.

\shortp{\python Code} The assistant must help the user write a function in the Python programming language. The original instructions were sourced from the HumanEval \cite{chen2021evaluating} and LiveCodeBench \cite{jain2024livecodebench} datasets, two popular benchmarks used to evaluate LLM programming aptitude.

\shortp{\database Database} The assistant is provided with the schema of an SQL database and a user query in natural language, and must produce an SQL query that retrieves the requested information from the database (a.k.a., text-to-SQL). The original instructions and databases were sourced from the popular Spider dataset \cite{yu2018spider}.

\shortp{\apis Actions} The assistant is provided with a set of API (Application Programming Interface) schemas, and a user instruction that requires API use, and must generate the programmatic API commands that match the user request. We sourced API schemas and user instructions from the Berkeley Function Calling Leaderboard (BFCL) \cite{2024bfcl}, a popular benchmark used to measure LLM ability at API function calling.

\shortp{\tmath Math} The assistant is provided with an elementary math word problem, and must perform a series of calculations using basic arithmetic operations to reach a numerical answer. We sourced problems from the GSM8K dataset \cite{cobbe2021training}.

\shortp{\datattext Data-to-text} The assistant is provided tabular data and several elements of related metadata, and must produce a caption (natural language sentence) describing the underlying data. We leverage the ToTTo \cite{parikh2020totto} dataset to formulate sharded instructions. 

\shortp{\summary Summary} The assistant receives a corpus of around twenty documents and a user query, and must generate a summary with citations that addresses the query based on the documents. We re-purpose the instructions from Summary of a Haystack \cite{laban2024summary}. The summary task is the only task we include that tests long-context capabilities, with instructions spanning several tens of thousands of tokens, which is known to deteriorate model performance \cite{huang2023embrace,karpinska2024one,kim2024fables}.

For each task, we reuse the metrics used in the original benchmarks. More specifically, the first four tasks (Code, Database, Actions, and Math) are evaluated for binary correctness, either by executing an answer attempt (code, SQL query), or validating semantic equivalence to a reference answer (API call, numerical answer). The last two tasks (Data-to-Text and Summary) are \textit{refinement tasks}, which get scored on a continuous range (0-100). Data-to-text uses the BLEU metric \cite{papineni2002bleu}, and Summary uses a custom LLM-as-a-judge metric (``Joint Score'') built to measure information coverage and attribution accuracy of the summary \cite{laban2024summary}. We map binary accuracy in the range of 0-100 (0 = failure, 100 = success) so that all tasks produce scores on a common scale, facilitating aggregation.

Appendix~\ref{app:task_implementation} lists implementation details of the sharding process for each task, including the sample selection process and any task-specific logic that was implemented to facilitate reproducibility. Even though we intended for the six selected tasks to be representative of a wide range of LLM use cases, we put effort into making the sharding process efficient and reproducible, as we see the process itself as a contribution of our work. We envision that future LLM evaluation practitioners can shard their own dataset artifacts to study LLM multi-turn behavior in more diverse and unique settings.

\subsection{Metric Selection} \label{sec:metrics}

LLMs employ a stochastic process to generate text. When setting LLM generation parameters to their default (\textit{e.g.}, T=1.0), LLMs generate many distinct responses for a fixed conversation state. We leverage this property to conduct repeated simulations for a given instruction and observe the variations that occur.
Each simulation yields a score $S_{i}$ ranging from 0-100 that assesses the level of success of the LLM in completing the task by the end of the simulation. Based on the set of scores $S = \{S_i\}_{i=1}^N$ obtained from running $N$ simulations for an instruction, we define three metrics: \textbf{averaged performance} ($\overline{P}$), \textbf{aptitude} ($A^{90}$), and \textbf{unreliability} ($U_{10}^{90}$):

\vspace{-1.0em}
\begin{center}
\begin{minipage}{0.2\textwidth}
\[
\overline{P} = \displaystyle\sum_{i=1}^N S_{i} \big/ N
\]
\end{minipage}
\hfill
\begin{minipage}{0.3\textwidth}
\[
A^{90} = \operatorname{percentile}_{90}(S)
\]
\end{minipage}
\hfill
\begin{minipage}{0.48\textwidth}
\[
U_{10}^{90} = \operatorname{percentile}_{90}(S) - \operatorname{percentile}_{10}(S).
\]
\end{minipage}
\end{center}
\vspace{-0.3em}

Average performance $\overline{P}$ is an unbiased estimate of a model's mean score on an instruction in a given simulation type. Aptitude $A^{90}$ is an estimate of a model's 90th percentile score on a given instruction, a best-case metric that estimates scores obtained in the top 10\% of simulations conducted. Unreliability is an interpercentile range estimate, between the 90th and 10th percentile estimates, measuring the gap between best-case and worst-case simulations, giving a sense of \textit{level of degradation} that occurs in response quality due to stochasticity in the LLM.

Each of the metrics is computed on a per-instruction basis and can be averaged across a corpus of instructions to obtain corpus-level metrics. In the rest of the paper, we refer to reliability and unreliability interchangeably, with reliability defined as $R_{10}^{90} = 100 - U_{10}^{90}$. We also simplify the notations to $A$ for aptitude and $U$ for unreliability, though the metrics can be generalized to other percentile thresholds (\textit{e.g.}, $A^{80}$ or $U_{5}^{95}$).

In Appendix~\ref{app:concrete_a_vs_r}, we go over a concrete example of how an average degradation in performance ($\overline{P}$) from 90\% to 60\% could be due to a loss in aptitude, reliability, or a combination. Finally, Figure~\ref{fig:reliability_vs_ability} visually connects the aptitude and unreliability metrics to score box-plot visualizations. In summary, the height of the upper whisker of the box plot represents aptitude (A), and the distance between the upper and lower whiskers of the plot represents Unreliability (U).

\section{Simulation Scale and Parameters} \label{sec:scale_and_parameters}

In the main simulation experiment, we leveraged the totality of instructions we sharded across six tasks (a total of 600 instructions), and simulated conversations across three types: \full\hspace{0.1em} \fullt, \concat\hspace{0.1em} \concatt, and \sharded\hspace{0.1em} \shardedt. We experimented with 15 LLMs, running $N=10$ simulations for each pair of model and simulation type, totaling more than 200,000 simulated conversations. All simulations were conducted with a default temperature of $T=1$, however, we conducted a supplementary experiment (Section~\ref{sec:implications_llms}) that explores the effect of temperature on aptitude and reliability.

Although simulating ten conversations for each \texttt{(LLM, instruction, simulation type)} increases experimental costs ten-fold, it allows us to not only measure averaged performance ($\overline{P}$) more accurately, but also study aptitude and reliability of LLM systems in depth in Section~\ref{sec:reliability}.

We selected a total of 15 LLMs from eight model families: \openai~OpenAI (GPT-4o-mini, GPT-4o \cite{hurst2024gpt}, o3 \cite{openai2025o3}, and GPT-4.1), \claude~Anthropic (Claude 3 Haiku, Claude 3.7 Sonnet), Google's \gemini~Gemini (Gemini 2.5 Flash, Gemini 2.5 Pro) \cite{team2023gemini}, Meta's \llama~Llama (Llama3.1-8B-Instruct, Llama3.3-70B-Instruct, Llama 4 Scout) \cite{grattafiori2024llama}, \aitwo~AI2 OLMo-2-13B \cite{olmo20242}, \microsoft~Microsoft Phi-4 \cite{abdin2024phi}, \deepseek~Deepseek-R1 \cite{guo2025deepseek}, and \cohere~Cohere Command-A \cite{cohere2025command}. This selection prioritizes the evaluation of state-of-the-art models, including both small (8B) and large models (300B+). We purposefully include both open- and closed-weights models, as well as two reasoning models (o3, R1) to study the effect additional thinking (test-time compute) has on multi-turn conversation capability. Details on model versioning and access are listed in Appendix~\ref{app:model_details}. We estimate the total cost of conducting simulations to be around \$5,000.

\section{Results} \label{sec:results}

\begin{table}[t]
    \centering
\renewcommand{\arraystretch}{1.6} 
\setlength{\tabcolsep}{4pt}
\resizebox{1.0\textwidth}{!}{%
\begin{tabular}{lcccccc/cccccc/cccccc/cc}
  & \multicolumn{18}{c}{\Large{Lost in Conversation Experiment}} & & \\
\cmidrule(r{7pt}){2-19} 
 \multirow{ 2}{*}{\large{Model}} & \multicolumn{6}{c}{\full~\fullt} & \multicolumn{6}{c}{\concat~\concatt} & \multicolumn{6}{c}{\sharded~\shardedt} & \multicolumn{2}{c}{\Large{\textbf{Overall}}} \\
\cmidrule(r{7pt}){2-7} \cmidrule(r{7pt}){8-13}  \cmidrule(r{7pt}){14-19} \cmidrule(){20-21} 
 & \python & \database & \apis & \datattext & \tmath & \summary & \python & \database & \apis & \datattext & \tmath & \summary & \python & \database & \apis & \datattext & \tmath & \summary & \ccol{EEEEEE} \raisebox{0.5ex}{\concat} \raisebox{-0.5ex}{\LARGE{/}} \raisebox{-0.5ex}{\full}  &  \ccol{EEEEEE} \raisebox{0.5ex}{\sharded} \raisebox{-0.5ex}{\LARGE{/}} \raisebox{-0.5ex}{\full} \\

\cmidrule(lr){1-1} \cmidrule(r{7pt}){2-7} \cmidrule(r{7pt}){8-13}  \cmidrule(r{7pt}){14-19} \cmidrule(){20-21} 

\llama~3.1-8B & \ccol{FFFFFF} 27.4 & \ccol{FFFFFF} 64.1 & \ccol{FFFFFF} 82.9 & \ccol{FFFFFF} 13.7 & \ccol{FFFFFF} 63.9 & \ccol{FFFFFF} \phantom{0}7.6 & \ccol{F0B9B9} 21.2 & \ccol{EEB0B0} 47.7 & \ccol{FEFEFE} 83.0 & \ccol{FEFEFE} 15.7 & \ccol{FDF8F8} 62.6 & \ccol{F5D0D0} \phantom{0}6.5 & \ccol{F1BFBF} 21.7 & \ccol{E58484} 25.9 & \ccol{E58484} 45.5 & \ccol{FDF5F5} 13.3 & \ccol{E58484} 37.4 & \ccol{E58484} \phantom{0}3.4 &  91.6 &  62.5 \\
\aitwo~OLMo2 & \ccol{FFFFFF} 18.8 & \ccol{FFFFFF} 54.8 & \ccol{FFFFFF} 56.1 & \ccol{FFFFFF} 17.2 & \ccol{FFFFFF} 80.0 &  - & \ccol{F6D6D6} 16.3 & \ccol{EEAFAF} 40.5 & \ccol{F7DCDC} 49.8 & \ccol{F4CBCB} 14.3 & \ccol{FEFEFE} 80.1 &  - & \ccol{F0B7B7} 14.4 & \ccol{E58484} 22.4 & \ccol{E58484} 13.8 & \ccol{E58484} \phantom{0}9.0 & \ccol{E58484} 46.3 &  - &  86.5 &  50.5 \\
\claude~3-Haiku & \ccol{FFFFFF} 44.8 & \ccol{FFFFFF} 85.0 & \ccol{FFFFFF} 83.5 & \ccol{FFFFFF} 29.8 & \ccol{FFFFFF} 73.9 & \ccol{FFFFFF} 11.6 & \ccol{F2C4C4} 36.3 & \ccol{F8E0E0} 76.5 & \ccol{FCF2F2} 80.2 & \ccol{FEFEFE} 30.1 & \ccol{FEFEFE} 76.1 & \ccol{F1BDBD} \phantom{0}9.2 & \ccol{ECA4A4} 31.5 & \ccol{E58484} 31.8 & \ccol{E99999} 55.9 & \ccol{E68B8B} 18.6 & \ccol{E78F8F} 47.1 & \ccol{E58484} \phantom{0}1.6 &  91.6 &  52.4 \\
\openai~4o-mini & \ccol{FFFFFF} 75.9 & \ccol{FFFFFF} 89.3 & \ccol{FFFFFF} 94.1 & \ccol{FFFFFF} 35.9 & \ccol{FFFFFF} 88.1 & \ccol{FFFFFF} 14.9 & \ccol{F7D9D9} 66.7 & \ccol{FEFEFE} 90.7 & \ccol{FDF8F8} 92.2 & \ccol{F6D7D7} 31.2 & \ccol{FEFEFE} 88.0 & \ccol{F4CCCC} 12.5 & \ccol{E99797} 50.3 & \ccol{E58484} 40.2 & \ccol{E58484} 52.4 & \ccol{E58484} 19.8 & \ccol{E99999} 58.7 & \ccol{E58484} \phantom{0}7.2 &  93.0 &  56.2 \\
\llama~3.3-70B & \ccol{FFFFFF} 72.0 & \ccol{FFFFFF} 91.1 & \ccol{FFFFFF} 95.0 & \ccol{FFFFFF} 34.1 & \ccol{FFFFFF} 91.7 & \ccol{FFFFFF} 15.8 & \ccol{EDACAC} 52.7 & \ccol{FCF4F4} 87.9 & \ccol{FEFEFE} 97.0 & \ccol{FBECEC} 32.0 & \ccol{FEFEFE} 91.8 & \ccol{FAEBEB} 14.7 & \ccol{ECA8A8} 51.6 & \ccol{E58484} 35.4 & \ccol{EEB1B1} 71.0 & \ccol{E99696} 22.4 & \ccol{E99A9A} 61.5 & \ccol{E99999} 10.5 &  93.2 &  64.2 \\
\microsoft~Phi-4  & \ccol{FFFFFF} 53.2 & \ccol{FFFFFF} 87.6 & \ccol{FFFFFF} 82.7 & \ccol{FFFFFF} 23.9 & \ccol{FFFFFF} 89.2 &  - & \ccol{F9E3E3} 48.4 & \ccol{F9E3E3} 79.6 & \ccol{F9E6E6} 76.0 & \ccol{FDFEFE} 28.6 & \ccol{FEFEFE} 90.4 &  - & \ccol{EEADAD} 39.1 & \ccol{E58484} 33.1 & \ccol{E58484} 34.1 & \ccol{FDF5F5} 23.2 & \ccol{E58484} 52.5 &  - &  99.0 &  61.7 \\
\cohere~CMD-A & \ccol{FFFFFF} 72.0 & \ccol{FFFFFF} 91.9 & \ccol{FFFFFF} 98.5 & \ccol{FFFFFF} 27.7 & \ccol{FFFFFF} 94.5 & \ccol{FFFFFF} 24.3 & \ccol{F5D2D2} 61.6 & \ccol{FAEBEB} 86.1 & \ccol{FEFEFE} 98.4 & \ccol{FDFEFE} 33.2 & \ccol{FDF6F6} 91.9 & \ccol{F6D8D8} 21.3 & \ccol{E78B8B} 44.9 & \ccol{E58484} 33.6 & \ccol{EDACAC} 72.0 & \ccol{FEFEFE} 27.9 & \ccol{EBA2A2} 66.0 & \ccol{E58484} \phantom{0}4.9 &  97.3 &  60.4 \\
\llama~4-Scout & \ccol{FFFFFF} 73.9 & \ccol{FFFFFF} 92.7 & \ccol{FFFFFF} 98.0 & \ccol{FFFFFF} 35.2 & \ccol{FFFFFF} 96.3 & \ccol{FFFFFF} 13.7 & \ccol{F3C6C6} 60.3 & \ccol{F7D9D9} 81.5 & \ccol{FEFEFE} 98.3 & \ccol{F2C1C1} 28.2 & \ccol{FCF4F4} 92.9 & \ccol{FEFEFE} 13.7 & \ccol{E78D8D} 46.4 & \ccol{E58484} 27.1 & \ccol{ECA7A7} 69.9 & \ccol{EEAFAF} 26.1 & \ccol{EBA1A1} 67.0 & \ccol{F8DFDF} 12.3 &  91.0 &  66.1 \\
\openai~o3 & \ccol{FFFFFF} 86.4 & \ccol{FFFFFF} 92.0 & \ccol{FFFFFF} 89.8 & \ccol{FFFFFF} 40.2 & \ccol{FFFFFF} 81.6 & \ccol{FFFFFF} 30.7 & \ccol{FEFEFE} 87.2 & \ccol{F8E2E2} 83.3 & \ccol{FEFEFE} 91.5 & \ccol{FDF9F9} 39.4 & \ccol{FDF9F9} 80.0 & \ccol{FEFBFB} 30.4 & \ccol{E68888} 53.0 & \ccol{E58484} 35.4 & \ccol{E99A9A} 60.2 & \ccol{E58484} 21.7 & \ccol{F0B9B9} 63.1 & \ccol{F6D5D5} 26.5 &  98.1 &  64.1 \\
\claude~3.7-Sonnet & \ccol{FFFFFF} 78.0 & \ccol{FFFFFF} 93.9 & \ccol{FFFFFF} 95.4 & \ccol{FFFFFF} 45.6 & \ccol{FFFFFF} 85.4 & \ccol{FFFFFF} 29.3 & \ccol{FDF7F7} 76.2 & \ccol{F6D6D6} 81.5 & \ccol{FEFEFE} 96.0 & \ccol{FDFEFE} 53.3 & \ccol{FEFEFE} 87.2 & \ccol{FEFAFA} 28.9 & \ccol{F4CECE} 65.6 & \ccol{E58484} 34.9 & \ccol{E58484} 33.3 & \ccol{F0B8B8} 35.1 & \ccol{F3C7C7} 70.0 & \ccol{F2C3C3} 23.6 &  100.4 &  65.9 \\
\deepseek~R1 & \ccol{FFFFFF} 99.4 & \ccol{FFFFFF} 92.1 & \ccol{FFFFFF} 97.0 & \ccol{FFFFFF} 27.0 & \ccol{FFFFFF} 95.5 & \ccol{FFFFFF} 26.1 & \ccol{FDF7F7} 97.1 & \ccol{FDF7F7} 89.9 & \ccol{FFFFFF} 97.0 & \ccol{FCFDFE} 36.7 & \ccol{FDF6F6} 92.9 & \ccol{FAEBEB} 24.4 & \ccol{ECA7A7} 70.9 & \ccol{E58484} 31.5 & \ccol{E58484} 47.5 & \ccol{EEAFAF} 20.0 & \ccol{ECA4A4} 67.3 & \ccol{E99696} 17.2 &  103.6 &  60.8 \\
\openai~4o & \ccol{FFFFFF} 88.4 & \ccol{FFFFFF} 93.6 & \ccol{FFFFFF} 96.1 & \ccol{FFFFFF} 42.1 & \ccol{FFFFFF} 93.8 & \ccol{FFFFFF} 23.9 & \ccol{FBEBEB} 82.9 & \ccol{FDF8F8} 91.7 & \ccol{FEFEFE} 97.1 & \ccol{EFB6B6} 32.2 & \ccol{FDF8F8} 91.9 & \ccol{FEFEFE} 23.9 & \ccol{EBA1A1} 61.3 & \ccol{E58484} 42.3 & \ccol{EA9B9B} 65.0 & \ccol{E58484} 20.5 & \ccol{EDAAAA} 67.9 & \ccol{E58484} 10.6 &  94.5 &  57.9 \\
\gemini~2.5-Flash & \ccol{FFFFFF} 97.0 & \ccol{FFFFFF} 96.3 & \ccol{FFFFFF} 88.4 & \ccol{FFFFFF} 51.2 & \ccol{FFFFFF} 90.6 & \ccol{FFFFFF} 29.1 & \ccol{FCF0F0} 92.5 & \ccol{FEFCFC} 95.5 & \ccol{FEFEFE} 89.2 & \ccol{FEFEFE} 51.9 & \ccol{FDF7F7} 88.4 & \ccol{FEFEFE} 29.4 & \ccol{ECA4A4} 68.3 & \ccol{E58484} 51.3 & \ccol{E58484} 42.6 & \ccol{E58686} 31.0 & \ccol{EDACAC} 66.1 & \ccol{F8DEDE} 26.1 &  99.3 &  65.8 \\
\openai~4.1 & \ccol{FFFFFF} 96.6 & \ccol{FFFFFF} 93.0 & \ccol{FFFFFF} 94.7 & \ccol{FFFFFF} 54.6 & \ccol{FFFFFF} 91.7 & \ccol{FFFFFF} 26.5 & \ccol{F9E5E5} 88.7 & \ccol{FAE9E9} 86.5 & \ccol{FEFEFE} 98.5 & \ccol{FEFDFD} 54.4 & \ccol{FDF8F8} 89.7 & \ccol{FEFEFE} 26.8 & \ccol{EFB2B2} 72.6 & \ccol{E58484} 46.0 & \ccol{E99898} 62.9 & \ccol{E58484} 28.6 & \ccol{F0B8B8} 70.7 & \ccol{E58484} 13.3 &  97.9 &  61.8 \\
\gemini~2.5-Pro & \ccol{FFFFFF} 97.4 & \ccol{FFFFFF} 97.3 & \ccol{FFFFFF} 97.8 & \ccol{FFFFFF} 54.8 & \ccol{FFFFFF} 90.2 & \ccol{FFFFFF} 31.2 & \ccol{FDF9F9} 95.7 & \ccol{FDF7F7} 94.9 & \ccol{FEFEFE} 98.1 & \ccol{FEFEFE} 56.9 & \ccol{FEFBFB} 89.3 & \ccol{FEFEFE} 31.8 & \ccol{EBA2A2} 68.1 & \ccol{E58484} 43.8 & \ccol{E58484} 36.3 & \ccol{F4CECE} 46.2 & \ccol{ECA7A7} 64.3 & \ccol{F2C0C0} 24.9 &  100.1 &  64.5 \\

\bottomrule
\end{tabular}
}
\vspace{1.0ex}

    \caption{Averaged Performance ($\overline{P}$) of LLMs on six tasks (\python~Code, \database~Database, \apis~Actions, \datattext~Data-to-text, \tmath~Math, and \summary~Summary). For each task, conversations are simulated in three settings: \full~\fullt, \concat~\concatt, and \sharded~\shardedt. \underline{Models are sorted in ascending order of average \fullt scores across tasks}. Background color indicates the level of \hl{degradation} from the \fullt setting. The last two columns average the performance drops from the \concatt and \shardedt compared to the \fullt in percentages across the six tasks.}
    \label{tab:main_results}
\vspace{-4.0ex}

\end{table}

\subsection{Average Performance Findings} \label{sec:main_results}

Table~\ref{tab:main_results} summarizes results from the simulation. At a high level, \textbf{every model sees its performance degrade on every task when comparing \fullt and \shardedt performance, with an average degradation of -39\%}. We name this phenomenon \texttt{Lost in Conversation}: models that achieve stellar (90\%+) performance in the lab-like setting of fully-specified, single-turn conversation struggle on \textit{the exact same tasks} in a more realistic setting when the conversation is underspecified and multi-turn.

In comparison, models perform roughly equivalently in the \concatt setting, with \concatt performance averaging 95.1\% of the \fullt counterpart.
This implies that the loss in performance for \shardedt is not explained by potential loss of information in sharded instructions, as such a loss would be reflected in lower \concatt performance. We observe that smaller models (Llama3.1-8B-Instruct, OLMo-2-13B, Claude 3 Haiku) have more pronounced \concatt degradations (86-92), and interpret this as indicating that smaller models struggle to generalize as well as larger models: benign rephrasing affects performance more than for larger, more robust models. This lack of robustness to paraphrasing can be observed visually in Table~\ref{tab:main_results}: \concatt degradation (red background) is more pronounced in the top rows (weaker models) than the bottom rows (stronger models).

The last column of the Table (\raisebox{0.5ex}{\sharded} \raisebox{-0.5ex}{\LARGE{/}} \raisebox{-0.5ex}{\full}) aggregates performance degradation across the six tasks, summarizing the magnitude of the Lost in Conversation effect for each model.
Surprisingly, \textbf{more performant models (Claude 3.7 Sonnet, Gemini 2.5, GPT-4.1) get equally lost in conversation compared to smaller models (Llama3.1-8B-Instruct, Phi-4)}, with average degradations of 30-40\%. This is in part due to metric definitions. Since smaller models achieve lower absolute scores in \fullt, they have less scope for degradation than the better models. In short, no matter how strong an LLM's single-turn performance is, we observe large performance degradations in the multi-turn setting.

When looking at the task-specific breakdown, some models see more muted degradations in certain tasks. For instance, Command-A sees the least degradation on the Actions task, while Claude 3.7 Sonnet and GPT-4.1 conserve performance well on Code, and Gemini 2.5 Pro in the Data-to-Text task. This finding indicates that the multi-turn capabilities of models are not uniform across domains and validates the importance of benchmarking models across a wide variety of tasks to investigate model capabilities.

Additional test-time compute (reasoning tokens) does not help models navigate multi-turn underspecification, as the two reasoning models included in the experiment (o3, Deepseek-R1) deteriorate in similar ways to non-reasoning models. This result confirms that \textbf{additional test-time compute does not, on its own, allow models to strategize over multi-turn conversation}. The analysis we conduct identifies a potential root cause: reasoning models tend to generate lengthier responses (on avg. 33\% longer than non-reasoning LLMs). As we find in Appendix~\ref{app:qualitative}, longer assistant responses tend to contain more assumptions, which can derail the conversation by confusing the model on what requirements were posed by the user vs. its own previous turn responses.

\subsection{Aptitude vs. Reliability Analysis}  \label{sec:reliability}

\begin{figure}[t]
    \centering
    \begin{subfigure}{0.2\textwidth}
        \centering
        \includegraphics[width=\textwidth]{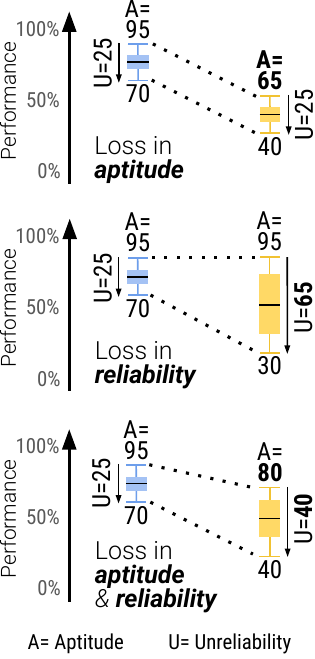}
        \caption{Visualizing Aptitude and Unreliability.}
        \label{fig:reliability_vs_ability}
    \end{subfigure}
    \hfill
    \begin{subfigure}{0.55\textwidth}
        \centering
        \includegraphics[width=\textwidth]{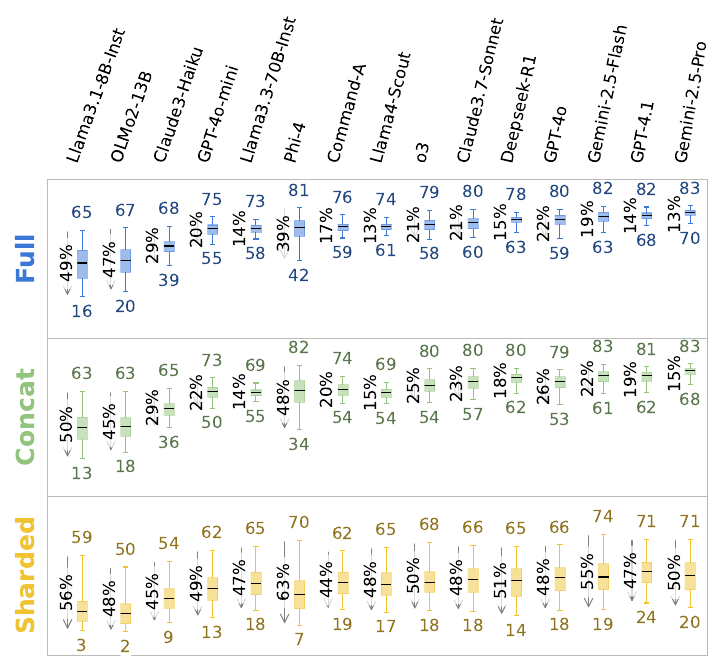}
        \caption{Observed Model Degradations}
        \label{fig:observed_boxplots}
    \end{subfigure}
    \hfill
    \begin{subfigure}{0.22\textwidth}
        \centering
        \includegraphics[width=\textwidth]{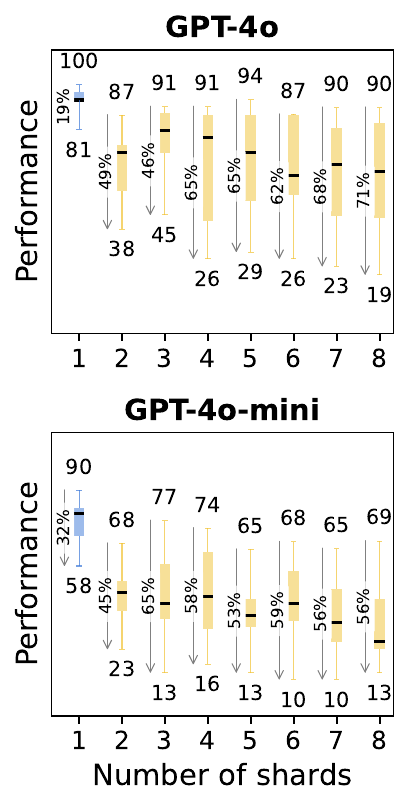}
        \caption{Gradual Sharding Results}
        \label{fig:gradual_sharding}
    \end{subfigure}
    \caption{(\subref{fig:reliability_vs_ability}) Visual introduction to the concepts of Aptitude and Unreliability when overlaid on a box-plot visualization, (\subref{fig:observed_boxplots}) reliability results based on experimental simulations with 15 LLMs, (\subref{fig:gradual_sharding}) summary of results from gradual sharding experiment, with instructions sharded in gradually larger shard sets (from 1 to 8 shards).}
    \label{fig:reliability}
\end{figure}

Results presented in Table~\ref{tab:main_results} present averaged performance degradation ($\overline{P}$). We now report on the aptitude and reliability analysis based on metrics $A$ and $U$. Figure~\ref{fig:observed_boxplots} visually summarizes the results of the reliability analysis we conducted on the 15 LLMs included in our simulation experiment. First, looking at the two single-turn settings, we see that models that are more able (higher A) tend to be more reliable (lower U). For instance, the two most able models (GPT-4.1 and Gemini 2.5 Pro) achieve the lowest unreliability. At the lower end, the two models with the lowest aptitude (Llama3.1-8B-Instruct and OLMo-2-13B) are also the most unreliable. In summary, \textbf{in single-turn settings, models with higher aptitude tend to be more reliable.} This fact is known in the community, with arguments made that better models require less prompt engineering, as they are more robust to minor variations in inputs and outputs \cite{li2023instruction}.

The sharded setting paints a different picture. Model aptitude degrades in a non-significant way between the full and sharded settings, with an average drop of 16\%. On the other hand, unreliability skyrockets with an average increase of 112\% (more than doubling). More interestingly, though better models tend to have slightly higher multi-turn aptitude, all models tend to have similar levels of unreliability. In other words, \textbf{in multi-turn, underspecified settings, all models we test exhibit very high unreliability, with performance degrading 50 percent points on average between the best and worst simulated run for a fixed instruction.} This refines our definition of the \textit{lost in conversation} phenomenon: when comparing single- and multi-turn settings, we find that large performance degradations ($\overline{P}$) are due in large part to increased model unreliability (U), rather than a loss in aptitude (A).

Appendix~\ref{app:qualitative} explores potential root causes for models getting lost in conversations. We identify four specific causes: (1) LLMs prematurely propose full answer attempts, making assumptions about problem specifications that lead to confusion (Appendix~\ref{app:premature_answer_attempt}), (2) they overly rely on previous (incorrect) answer attempts leading to lengthier ``bloated'' answers (Section~\ref{app:answer_bloat}), (3) LLMs overly adjust their answers based on the first and last turn of conversation, evidenced by a loss-of-middle-turns phenomenon (Appendix~\ref{app:summary_last_turn}), and (4) they produce overly verbose answers, which likely introduces assumptions that detract attention from user utterances (Section~\ref{app:assistant_verbosity}).

\subsection{Gradual Sharding Experiment} \label{sec:gradual_sharding}

The multi-turn conversations simulated based on sharded conversations are not representative of underspecified conversations that users might have with LLMs in realistic settings. In particular, the fact that sharded instructions must be maximal (property P3) and that the simulated user must reveal at most one shard of information per turn (Section~\ref{sec:simulation_process}) can seem unrealistic and adversarial. In fact, prior work has found that minor and severe underspecification appear in equal proportions in public LLM chat logs \cite{herlihy2024overcoming}. To explore the relationship between the granularity of sharding and the lost in conversation phenomenon, we propose the gradual sharding experiment.

In the gradual sharding experiment, we selected 31 instructions from our original experiment across multiple tasks, and expanded each sharded instruction into seven sharded instructions, with the shard-set size growing from 2 to 8 shards. The instruction selection and sharding process are detailed in Appendix~\ref{app:gradual_sharding}. The process ensured that at each shard set size (from 1 to 8), task complexity is fixed, and the only modified factor is the granularity of sharding.

We ran simulations for the gradual sharding experiments with two models (GPT-4o and GPT-4o-mini), with results summarized in Figure~\ref{fig:gradual_sharding}. We find that both models get lost in conversation (a minor degradation in aptitude and a large increase in unreliability) with two-shard instructions and beyond. In other words, the gradual sharding experiment indicates that \textbf{any conversation that involves underspecification and occurs in two or more turns leads to models getting lost in conversation}. For users, the granularity at which information is specified does not majorly impact reliability: providing all the information at once (1-shard) is the only effective method to improve reliability.

\section{Implications} \label{sec:implications}

\subsection{Implications for System and Agent Builders} \label{sec:implications_systems}

\begin{wraptable}{l}{0.34\textwidth}
    \centering
    \vspace{-1.0em} 
    \resizebox{1.0\linewidth}{!}{%
    \begin{tabular}{lccccc}
        & \multicolumn{5}{c}{Simulation Type} \\
        \cmidrule(){2-6}
        Model & \full & \concat & \sharded & \recap & \snowball \\
        \cmidrule(lr){1-1} \cmidrule(){2-6}
        \openai~4o-mini &  86.8 & \ccol{FDF6F6} 84.4 & \ccol{E58484} 50.4 & \ccol{F0B7B7} 66.5 & \ccol{ECA6A6} 61.8 \\
        \openai~4o &  93.0 & \ccol{FDF8F8} 90.9 & \ccol{E78F8F} 59.1 & \ccol{F3C9C9} 76.6 & \ccol{EBA3A3} 65.3 \\
        \bottomrule
    \end{tabular}
    }
    \caption{Experimental Results with additional simulation types: \recap\hspace{0.1em} Recap and \snowball\hspace{0.1em} Snowball. Both strategies involve repeating user-turn information to mitigate models getting lost in conversations.}
    \label{tab:recap_snowball}
    \vspace{-4ex}
\end{wraptable}

Building LLM-based applications typically involves complex processes: decomposition of problems, retrieval of relevant information, use of tools, and calling of actions. Such processes are typically orchestrated by an agentic framework (such as Autogen \cite{wu2023autogen} or LangChain \cite{chase2022langchain}) that allows system builders to compose workflows with LLM calls as individual blocks. As such, an argument could be made that multi-turn capabilities are not a necessary feature of LLMs, as it can be offloaded to the agent framework. In other words, do we need native multi-turn support in LLMs when an agent framework can orchestrate interactions with users and leverage LLMs only as single-turn operators?

To answer this question, we implemented two agent-style conversation simulation types: \recap~\recapt and \snowball~\snowballt. Both preprocess user utterances before sending them to the LLM. In \recapt, a conversation proceeds in the same way as \shardedt, but a user turn is added at the end, which recapitulates all the previous user turns. \snowballt is a more gradual recapitulation: at each turn, the user simulator reveals a new shard, and repeats all previously revealed shards at that point. Both simulation types repeat the past user's turn information to make it more prominent and give the LLM a chance to leverage the redundancy to improve its responses. We include the experimental detail in Appendix~\ref{app:recap_snowball}.

Table~\ref{tab:recap_snowball} summarizes the results on all instructions for four tasks (Code, Database, Math, Actions) for two tested models (GPT-4o, GPT-4o-mini). Both \recapt and \snowballt demonstrate some level of success, with improvements over \shardedt simulations, but the performance still lags behind \fullt or \concatt. While \recapt outperforms \snowballt, we note that \recapt is an unrealistic setting because the intervention is conducted on the \textit{last turn} of the conversation, which is not known a priori when conversation unfolds with a real user. \snowballt gives a sense of realistic performance gains achievable through user-turn repetition: it can mitigate the \fullt-to-\shardedt performance deterioration by 15-20\%. In short, relying on an agent-like framework to process information might be limiting, and we argue LLMs should natively support multi-turn interaction.

\subsection{Implications for LLM Builders} \label{sec:implications_llms}

A lot of effort has been put in improving LLM \textit{aptitude}: demonstrating that LLMs can accomplish tasks of increasing intellectual complexity, with recent results showing LLMs can compete in mathematics Olympiads, or solve Ph.D.-level technical questions in a benchmark aptly named Humanity's Last Exam \cite{phan2025humanity}.

In this work, we call on LLM builders to prioritize \textit{reliability} of the models they build, as our experiments demonstrate that the randomness involved in generating text with LLMs leads to catastrophic unreliability in all the models we tested, degrading the quality of responses the average LLM users see.

LLMs are probabilistic systems, with parameters such as \textit{temperature} that can adjust the degree of randomness that occurs while generating text. A possible argument is therefore: does setting the temperature to its lowest setting ($T=0$) effectively resolve the reliability concern, as it makes the generation process more (but not entirely) deterministic?

To evaluate this argument, we conducted a supplementary experiment in which the assistant's temperature for generating responses (AT) was varied to three values: 1.0, 0.5, and 0.0. Additionally, since \shardedt simulation uses an LLM-based user simulator, we also varied the user's temperature (UT) with the same three values. Further details on the experiment, including sample selection and simulation scale, are in Appendix~\ref{app:temperature}.

\begin{wraptable}{r}{0.4\textwidth}
    \centering
    \vspace{-0.5em} 
    \setlength{\tabcolsep}{4pt}
    \resizebox{1.0\linewidth}{!}{%
    \begin{tabular}{lccc/ccc}

    & \multicolumn{3}{c}{\openai~4o-mini} & \multicolumn{3}{c}{\openai~4o} \\
     \cmidrule(lr){2-4} \cmidrule(lr){5-7}
    Simulation & \small{AT=1.0} & \small{AT=0.5} & \small{AT=0.0} & \small{AT=1.0} & \small{AT=0.5} & \small{AT=0.0} \\
    
    \cmidrule(lr){1-1} \cmidrule(r{7pt}){2-4} \cmidrule(){5-7}

\full\hspace{0.1em}  \fullt & \ccol{F7D9D9}16.0 & \ccol{F7DBDB}15.0 & \ccol{FBEFEF}6.8 & \ccol{F6D5D5}17.8 & \ccol{FBECEC}8.0 & \ccol{FDF8F8}2.8 \\
\concat\hspace{0.1em} \concatt  & \ccol{F5CFCF}20.2 & \ccol{F6D5D5}17.8 & \ccol{FAE8E8}9.5 & \ccol{F5CFCF}20.2 & \ccol{F6D5D5}17.8 & \ccol{FCF1F1}5.8 \\

    \cmidrule(lr){1-1} \cmidrule(r{7pt}){2-4} \cmidrule(){5-7}

\sharded\hspace{0.1em} \small{UT=1.0} & \ccol{E68989}49.8 & \ccol{E89090}46.8 & \ccol{E58686}51.0 & \ccol{EA9E9E}41.0 & \ccol{E99898}43.8 & \ccol{EFB4B4}31.8 \\
\sharded\hspace{0.1em} \small{UT=0.5} & \ccol{EFB4B4}31.7 & \ccol{EEAEAE}34.0 & \ccol{EB9F9F}40.5 & \ccol{EBA2A2}39.5 & \ccol{EB9F9F}40.8 & \ccol{EFB4B4}31.8 \\
\sharded\hspace{0.1em} \small{UT=0.0} & \ccol{ECA4A4}38.5 & \ccol{F1BDBD}28.0 & \ccol{F0B7B7}30.5 & \ccol{EDAAAA}35.8 & \ccol{ECA5A5}38.0 & \ccol{F0B9B9}29.7 \\

    \bottomrule
    \end{tabular}
    }
    \caption{Unreliability of models when changing assistant temperature (AT) and user temperature (UT) in \full~\fullt, \concat~\concatt and \sharded~\shardedt settings. The lower the number the more reliable the assistant is.}
    \label{tab:temperature}
    \vspace{-2ex}
\end{wraptable}

Table~\ref{tab:temperature} summarizes the experimental findings. Looking at the \fullt and \concatt settings (first two rows), both GPT-4o-mini and GPT-4o observe a large improvement in reliability when temperature is decreased, with a drop in unreliability ($U_{10}^{90}$) of 50-80\% when the assistant temperature decreases. Results from \shardedt simulations are more alarming: GPT-4o-mini does not see improvements in reliability as AT is decreased (in all user-temperature settings), and GPT-4o only sees minor improvements, on the order of 15-20\%. Even when both the user and assistant temperatures are set to 0.0, there remains a large unreliability of around 30\%. Even though language models are supposed to be deterministic at $T=0.0$, this is known to practically not be the case for modern LLMs (see Appendix~\ref{app:randomness} for discussion).
At a high level, single-turn conversations have limited scope for deviation, whereas one token difference in an early turn of a multi-turn conversation can lead to cascading deviations, which we observe as stagnated unreliability.
For settings that involve multi-turn interaction, we find that \textbf{lowering the temperature of the LLM when generating responses is ineffective in improving system reliability.}

We invite and challenge LLM builders to jointly optimize model aptitude and reliability. A reliable LLM should: (1) achieve similar aptitude in single- and multi-turn settings, (2) have small unreliability ($U_{10}^{90} < 15$) in multi-turn settings, (3) achieve these at unmodified temperature ($T=1.0$), demonstrating that the underlying language model can handle variations that naturally occur in language generation.

\subsection{Implications for NLP Practitioners} \label{sec:implications_nlp}

Our experiments demonstrate that model behavior in single- and multi-turn settings on the same underlying set of instructions can diverge in important ways, for example, with large observed degradations in performance and reliability.

We selected the initial six tasks to span a wide range of generation tasks, from programming to multi-document summarization. Yet this set of tasks is limited across multiple dimensions, such as focusing on English-language instructions and analytical (i.e., non-creative) tasks. We put effort into making the sharding process scalable by automating portions that could be handled by an LLM, while manually validating and finalizing samples for quality control. The sharding process -- detailed in Appendix~\ref{app:sharding_process} -- required an average of three hours of manual work (prompt engineering or inspection) from an author to prepare and finalize 100 sharded instructions.

We encourage NLP practitioners to experiment with sharding and release sharded versions of their tasks and instructions alongside fully specified ones.

\begin{wraptable}{l}{0.28\textwidth}
    \vspace{-2ex}
    \centering
    \resizebox{1.0\linewidth}{!}{%
    \begin{tabular}{lccc}
    & \multicolumn{3}{c}{\translation Translation}\\
     \cmidrule(){2-4}
    Model &\full&\concat& \sharded \\
    \cmidrule(lr){1-1} \cmidrule(){2-4}
    \openai~4o-mini &  41.7 &  43.4 &  42.1 \\
    \openai~4o &  35.9 &  38.5 &  40.9 \\
    \bottomrule
    \end{tabular}
    }
    \caption{Performance on the \translation translation task for \full~\fullt, \concat ~\concatt, and \sharded~\shardedt simulations.}
    \label{tab:translation}
    \vspace{-2ex}

\end{wraptable}

To illustrate the feasibility of sharding new tasks, and understand compatibility requirements for sharding, we prepared sharded instructions for a seventh task: \translation Translation. The task consists of translating an entire document (10 sentences) from German to English, leveraging paired documents from WMT 2019 on document-level translation \cite{scherrer2019wmt}. In the \shardedt setting, each turn reveals two additional sentences from the source document and requires the assistant to translate all sentences provided so far, whereas the \fullt and \concatt settings reveal the entire document in the first turn. Evaluation is conducted with the standard BLEU metric \cite{papineni2002bleu}. We describe practical implementation details in Appendix~\ref{app:task_implementation}.

Results from \fullt, \concatt, and \shardedt simulations are summarized in Table~\ref{tab:translation}. Both models we tested -- GPT-4o-mini and GPT-4o -- do \textit{not} exhibit degradation in performance in the \shardedt setting, with BLEU scores being within 10\% difference of each other in all settings. We believe this result reflects that the task can largely be accomplished at the sentence-level despite some prior work has framed translation at the document-level \cite{post2023escaping}, and that the BLEU score does not adequately capture document-level nuances \cite{ma2021comparison}.
In other words, if a task is episodic (i.e., it can be decomposed into turn-level subtasks), the models can avoid getting lost in conversation by completing each subtask without having to handle multi-turn context. In short, the \shardedt Translation task simulates multi-turn conversations that are not underspecified.

We now list task properties we believe are important in leading models to get lost in conversation in multi-turn settings. First, generative tasks (\textit{i.e.}, unlike extractive QA or classification) are more prone to model confusion, as they typically involve editing and refinement of new content. Second, the generative tasks should be sufficiently complex, involving multiple explicit specifications that will yield a multitude of shards. For example, an instruction: ``Write a Python program that calculates $1+1$'' is too simple to shard. Third, the solution or answer should be non-decomposable, such that revealing a shard modifies the entire solution (unlike the translation task, where each additional shard only asks to translate and append to the ongoing solution). We hypothesize that LLMs tested on tasks with the aforementioned three properties will likely get lost in conversation, evidenced by a large drop in averaged performance and reliability in \shardedt simulations.

\subsection{Implications for Users of Conversational Systems} \label{sec:implications_users}

Users of LLM-based products should be aware of the lack of reliability of LLMs, particularly when used in multi-turn settings. Generally available generative technology is new, and prior work has identified the randomness in LLM-generated text as a point of confusion for users \cite{Mylrea2023ArtificialI,Weisz2024DesignPF,venkit2024search,Lee2024OneVM}. We make two practical recommendations that can help users of LLM-based systems get the most out of their exchanges.

\paragraph{If time allows, try again.} If a conversation with an LLM did not lead to expected outcomes, starting a new conversation that repeats the same information might yield significantly better outcomes than continuing an ongoing conversation. This is because current LLMs can get lost in the conversation, and our experiments show that persisting in a conversation with the model is ineffective. In addition, since LLMs generate text with randomness, a new conversation may lead to improved outcomes.

\paragraph{Consolidate before retrying.} Since LLMs are ineffective at dealing with information dispersed across multiple turns, consolidating instruction requirements into a single instruction is an effective strategy to improve the model's aptitude and reliability (as shown by the \concatt experiments).
When a user notices that a model is lost in conversation, they can ask the LLM: ``Please consolidate everything I've told you so far,'' then bring the response to a new conversation, alleviating the need for manual consolidation. In practice, there is anecdotal evidence that early adopters of LLM-based applications are aware that LLMs get lost in conversation. For example, users of the Cursor LLM-based coding environment report that frequently creating new conversations ``whenever they can'' is a recommended strategy to ensure high quality responses even though the tool allows to keep conversations going indefinitely.\footnote{\url{https://www.reddit.com/r/cursor/comments/1j72r8d/when_to_start_a_new_chat/}} 

These two recommendations remain cumbersome for users and can only offer patched solutions rather than a principled approach. Once future LLMs can more reliably handle multi-turn conversations, the need for such recommendations should be alleviated, allowing users to communicate underspecified instructions over multiple turns naturally with less risk of the model getting lost in conversation.

\section{Conclusion}\label{sec:conclusion}

In this work, we conduct a large-scale simulation of single- and multi-turn conversations with LLMs, and find that on a fixed set of tasks, LLM performance degrades significantly in multi-turn, underspecified settings. LLMs get lost in conversation, which materializes as a significant decrease in reliability as models struggle to maintain context across turns, make premature assumptions, and over-rely on their previous responses. Additional experiments reveal that known remediations that work for simpler settings (such as agent-like concatenation or decreasing temperature during generation) are ineffective in multi-turn settings, and we call on LLM builders to prioritize the reliability of models in multi-turn settings.


\section{Limitations} \label{sec:limitations}

A first limitation of our work is the reliance on fully automated simulation. By relying on an LLM to simulate user utterances, we can scale our experiments, including running the same simulation multiple times, which would be cost-prohibitive with real users. However, the simulations we obtain are not representative of natural human-AI conversation. The properties of the sharding process (defined in Appendix~\ref{app:sharding_process}) and of the simulation environment (see Section~\ref{sec:simulation_process}) ensure that the simulated conversations follow a rather narrow structure, likely not modeling the full range of conversation dynamics that occur with a large, diverse user population. For example, the simulation process ensures a new shard of information is revealed at each turn, and that the last turn of the conversation has specified all the information needed to complete the task which might not happen with real users. Properties P1, P2, and P5 of the sharding process also restrict the scope of the conversation, as sharded instructions closely match an existing fully-specified instruction, with the high-level intent always identified in the conversation's first turn. The minimal nature of shards is also unrealistic and potentially adversarial, though the gradual sharding experiment finds that different levels of shard granularity lead to similar performance degradations, as soon as conversations occur over two turns or more.
Apart from sharding granularity, automatic simulation also lacks the nuance that can occur when a human is involved in conversation, from misunderstandings over terminology, giving up due to frustration with system failures \cite{wester2024ai}, or the lack of a feasible end goal for certain conversations (e.g., the user wanting a solution to an unsolved problem). Because of these factors, we believe conducted simulations represent a benign testing ground for LLM multi-turn capabilities. \textbf{Because of the overly simplified conditions of simulation, we believe the degradation observed in experiments is most likely an underestimate of LLM unreliability, and how frequently LLMs get lost in conversation in real-world settings.} The experiments serve as a scalable, low-cost experimental environment for studying LLMs in multi-turn settings.

A second limitation of our work is the focus on analytical tasks. Although we selected a diverse set of both programming and natural language tasks, we restricted experiments to tasks that involve an analytical solution. This restriction limits the scope of our findings, as we do not establish whether models get lost in conversation on more open-ended tasks, such as creative writing \cite{chakrabarty2024art}. This was a conscious choice: though there has been some progress on creative writing evaluation, it is still an active area of research \cite{chakrabarty2025ai}, and we relied on more established tasks and metrics for the initial set of experiments. Determining whether degradation occurs -- and if so, identifying the magnitude -- on creative tasks is an important direction for future work.

A third limitation of the work is the focus on text-only tasks in the English language. Establishing whether models get lost in conversation in other languages, or in tasks that involve multiple modalities in either user or assistant utterances, could help establish the scope of the degradation observed in LLM multi-turn capabilities.






\bigskip

\bibliography{bibliography}

\clearpage
\begin{appendices}

\part*{Appendices}

\section{Related work on Underspecification} \label{app:rel_work}

The Background (Section~\ref{sec:background}) reviews the most directly related prior work, focused on multi-turn evaluation. We now cover other related prior works that have studied underspecification.


Prior work on communication and linguistics has identified underspecification as a common feature of human language \cite{Lappin2000AnIP,ferreira2008ambiguity,Frisson2009SemanticUI,pezzelle2023dealing}.

Understanding how LLMs handle underspecified instructions is crucial towards improving conversational capabilities.
To this end,~\citet{herlihy2024overcoming} identified common response patterns such as hedging, refusal, clarification, and interrogation when underspecified queries are presented to conversational LLM systems, and proposed mechanisms to recover from them.
\citet{malaviya2024contextualized} highlighted the importance of supporting context for more accurate and principled evaluation of LLM responses on underspecified queries, and \citet{Sarkar2025ConversationalUI} showed that a system that proactively rewrites user instructions to account for underspecification leads to improved LLM response. \citet{shaikh2025navigating} studied the degree of grounding (\textit{i.e.}, clarifications and follow-up questions) that LLMs perform in conversation logs and observed that they significantly lack in generating follow-up questions, where humans are 15 times more likely to do so. \citet{chang2025chatbench} hired annotators to manually reproduce fully-specified instructions through a chat interface, and found that the users reveal the entirety of the instruction in 34\% of the time, leaving some detail underspecified a majority of the time.

Several works have explored direct tasks to evaluate model ability when dealing with underspecification. \citet{liu2023we} introduced AmbiEnt, a natural language inference benchmark, which revealed that understanding ambiguous statements is still a challenge even to the state-of-the-art LLMs. \citet{Wildenburg2024DoPL} created the DUST task, which requires the language model to determine the relative levels of specifications between two sentences, finding that when interpreting underspecified sentences, LMs exhibit little uncertainty.
\citet{vijayvargiya2025interactive} evaluated LLM agents for GitHub issue resolution in an underspecified setting, showing that follow-up interactions to supplement information helps improve the resolve rate but detecting the ambiguities in the instructions remains a challenge.

Prior work has classified different root causes for underspecification. First, task underspecification occurs when humans provide incomplete descriptions of the task at hand, which is prominent in ``specification-heavy tasks'' \cite{Peng2023WhenDI}. Second, intent misalignment occur when the AI fails to understand the user's intent or motivation, and is one of the common sources of user dissatisfaction \cite{Kim2024BeyondPL,Terry2023InteractiveAA}. Finally, \citet{Chaturvedi2024NeBuLaAD} discuss location and and reference ambiguity, in emboddied settings that involve physical spaces such as a Minecraft game.



\section{Precise Definition of Sharded Instructions} \label{app:sharded_definition}

Section~\ref{ssec:sharding_concept} introduces the concept of sharding at a high level. This Appendix offers a more precise definition by first defining mathematical terminology, and then defining properties that a sharded instruction must satisfy to be considered valid.

Let $q$ refer to a single-turn complex query with intended (i.e., correct) output $Y_q^*$. We refer to the atomic content units (ACU) \cite{liu2022revisiting} of the query as 
\[I(q)=[\mathcal{I},({c}_1, \cdots, c_m )]\]
where $\mathcal{I}$ is the primary intent of the query and $(c_1, \cdots, c_m )$ are the sufficient set of clarifications that specify details of how to compute $Y_q^*$ conditioned on $\mathcal{I}$. For $I(q)$ to be considered {\em atomic}, any rephrasing of $I(q)$ should produce the same target output. Ie. for all $q'$ s.t. $I(q')=I(q)$, then $Y_{q}'^* = Y_q^*$. 

Given the above definition, the {\em aim} of the sharding process, for a given query $q$, is to identify the atomic content units $I(q)$ and construct a set of shorter instruction {\em shards} $\mathbf{s}$:
\[ q' = [s_1, \cdots s_k] \; \mbox{ s.t. } I(q)=I(q')\]
where the shards $s_j$ can be used to simulate multi-turn conversation, with the same intended output as $q$.

A sharded instruction $q'$ is considered valid for an original query $q$ if it fulfills the following properties:

\shortp{P1: Information Preservation.} $I(q)=I(q')$ No information from the original instruction necessary for the completion of the instruction should be lost during the sharding process.
\shortp{P2: Clear Initial Intent.} $\mathcal{I}_q=\mathcal{I}_{q'}$ and $s_1=\mathcal{I}_q$. The first shard plays a distinctive role of being the \textit{initial query} within the shard set. The initial query defines the high-level objective for the entire conversation. (e.g., ``write a Python function'').
\shortp{P3: Order Insensitive.} Apart from the first shard, the other shards should be decontextualized \cite{choi2021decontextualization} and not refer to each other in a way that implies an order. As a result, the shard set presented in any order reveals equivalent information. Let $\rho(\mathbf{s}_{2..k})$ refer to a permutation of the shard ordering, then $I(q)=I(\tilde{q}) \; \forall \tilde{q}=[s_1,\rho(\mathbf{s}_{2..k})]$
\shortp{P4: Maximal Sharding.} The sharding process should strive to maximize the number of shards extracted from the original instruction (maximize $k$). This can be achieved by producing shards that introduce a single, specific piece of information.
\shortp{P5: Minimal Transformation.} The sharded instruction should maintain the instruction language and avoid simplifying, altering, or interpreting elements of the original instruction as much as possible. Apart from modifications to satisfy properties P1-P4, the sharding process should attempt to limit modifications such that the shards ($[s_1, \cdots s_k]$ are semantically similar to the atomic content units $I(q)$.


\section{Semi-Automatic Sharding Process} \label{app:sharding_process}

\begin{figure}[tb]
    \centering
    \includegraphics[width=\linewidth]{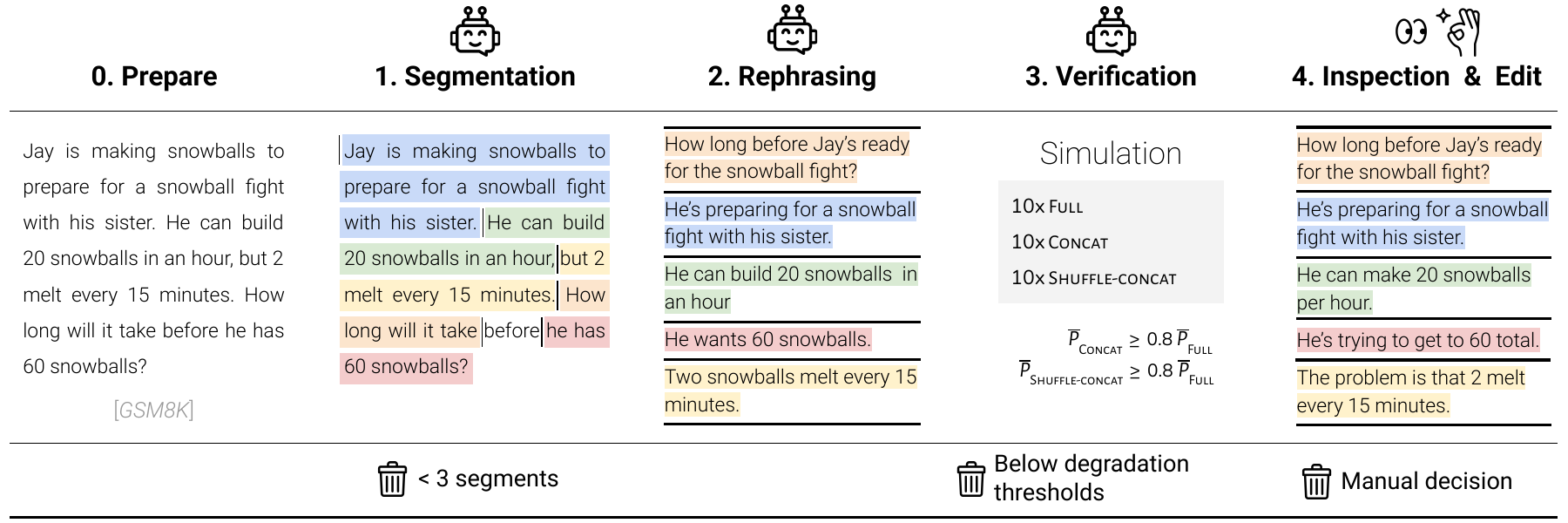}
    \caption{Process diagram of the four-step semi-automatic process to transform fully-specified instructions into a sharded instruction. The first three steps (segmentation, rephrasing, verification) are automated, while the fourth (inspect and edit) was manually completed by the authors of the work. The last row represents the rejection criteria for a sample.}
    \label{fig:sharding_process}
\end{figure}

We rely on a semi-automatic process to transform fully-specified instructions into their sharded equivalents. The process -- illustrated in Figure~\ref{fig:sharding_process} -- consists of a sequence of three automated steps (Segmentation, Rephrasing, Verification) followed by a manual step that was conducted by an author of the paper.

We now detail each step of the process, then go over task-specific details we implemented as needed. We note that as part of our open-source release, we provide all the prompts used in the first three LLM-based steps.

\paragraph{Step 1: Segmentation} Given an original fully-specified instruction (left-most column in Figure~\ref{fig:sharding_process}), the LLM is prompted to extract \textit{segments} of the instructions. Segments are intended to correspond to the atomic content units (defined in Appendix~\ref{app:sharded_definition}). In particular, the prompt indicates that segments must not overlap, and that not all words in the original instruction must belong to a segment. Prompts are task-specific and incorporate at least three few-shot examples of segmentation, to allow for the concept of segmentation to be illustrated through examples. At this stage, any instruction that yields fewer than three segments are filtered out and does not proceed to the next stage.

\paragraph{Step 2: Rephrasing} Given the original fully-specified instruction and the extracted segments, this stage consists in rewriting each segment to be decontextualized \cite{choi2021decontextualization} and conversational. In other words, dependencies between segments are resolved, and the ordering is changed such that obtained \textit{shards} adhere to properties P2 and P5. In the example above, the fourth segment (highlighted in orange) becomes the first shard as it reveals the overall intent, and light rephrasing occur in other shards. The rephrasing prompt is task-specific, and includes few-shot examples of rephrasing segmented instructions.

\paragraph{Step 3: Verification} Steps 1-2 produce a sharded instruction that can be used to simulate \shardedt and \concatt conversations. To verify the property P1 (Information Preservation) that no information has been lost during segmentation and rephrasing, we conduct preliminary simulations to evaluate the original and sharded instruction side-by-side. Specifically for each pair of the original and the sharded instruction, we simulate ten \fullt conversations with the original instruction, ten \concatt conversations with the sharded instruction (by concatenating the shards), and ten \textsc{Shuffle-concat} conversations. \textsc{Shuffle-concat} is a variant of the \concatt simulation in which all shards (except Shard 1) are randomly permuted before being concatenated. This variant can be seen as a more adversarial version of \concatt, verifying the property P3 (Order Insensitive). For each simulation type, we calculate the averaged performance $\overline{P}$ over ten runs and filter out instructions that are below an acceptable degradation threshold. Specifically, instructions are acceptable if the following conditions are met:
\begin{align*}
  \overline{P}_{\textsc{Concat}} &\geq 0.8~\overline{P}_{\textsc{Full}}\\
  \overline{P}_{\textsc{Shuffle-concat}} &\geq 0.8~\overline{P}_{\textsc{Full}},
\end{align*}
where $\overline{P}_{\textsc{X}}$ denotes the averaged performance of the simulation type \textsc{X}. If more degradation is observed (\textit{i.e.}, below 80\%), it indicates a potential loss of information during sharding, or that decontextualization was not implemented accurately.

\paragraph{Step 4: Inspect and Edit} Even though the first three steps define the sharding process and implement some level of quality assurance, they do not guarantee the level of quality required for precise and large-scale experiments due to relying on LLM outputs. To obtain high-quality shards, we reserve step 4 for manual inspection and validation. To facilitate the procedure, we developed a web-based annotation interface. In the interface, an annotator can review a pair of fully-specified and sharded instructions, edit, add, or remove individual shards, and decide to accept or reject sharded instructions. Sharded instructions included in our experiments were all manually reviewed by two authors of the work. The amount of editing and filtering required in this final stage varied by task.

Inspecting and editing an auto-generated instruction typically requires 1-3 minutes per instruction, an order of magnitude less than it would require for authors to write the sharded instructions de-novo from a given fully-specified instruction.
As part of our open-source release, we provide all the prompts used during sharding, which we hope can facilitate the sharding of additional tasks.

\section{Inspection of Simulated Sharded Conversation} \label{app:inspection}

\begin{wraptable}{l}{0.5\linewidth}
    \vspace{-1em}
    \centering
    \resizebox{1.0\linewidth}{!}{%
    \begin{tabular}{lccccc}
    Inspection & All Tasks & \apis Actions & \python Code & \tmath Math & \database Db \\

    \cmidrule(lr){1-1} \cmidrule(){2-6}

    Shard Fully Revealed & \ccol{FCE9E9}96.0 & \ccol{FDF5F5}98.3 & \ccol{FBE3E3}94.9 & \ccol{FADBDB}93.4 & \ccol{FFFFFF}100.0 \\
    Shard Contextualized & \ccol{FDF6F6}98.4 & \ccol{FDF5F5}98.3 & \ccol{FDF5F5}98.3 & \ccol{FDF6F6}98.3 & \ccol{FEF7F7}98.6 \\

    \cmidrule(lr){1-1} \cmidrule(){2-6}

    Strategy Accuracy & \ccol{FBE5E5}95.2 & \ccol{FBE2E2}94.7 & \ccol{FBE6E6}95.5 & \ccol{FBE7E7}95.6 & \ccol{FBE2E2}94.7 \\
    Extraction Success & \ccol{FCEEEE}97.0 & \ccol{FFFFFF}100.0 & \ccol{FADBDB}93.4 & \ccol{FDF6F6}98.4 & \ccol{FFFFFF}100.0 \\

    \cmidrule(lr){1-1} \cmidrule(){2-6}
    
    Overall Success & \ccol{FDF2F2}97.8 & \ccol{FFFFFF}100.0 & \ccol{FCE9E9}96.0 & \ccol{FCE9E9}96.0 & \ccol{FFFFFF}100.0 \\
    \bottomrule
    \end{tabular}
    }
    \caption{Results of the inspection of 100 simulated sharded conversations across four tasks: Actions, Code, Math, and Database. The first column aggregates annotation results on the four tasks.}
    \label{tab:inspection}
    \vspace{-1em}
\end{wraptable}

The sharding simulation environment (described in Section~\ref{sec:simulation}) relies on LLM components to simulate the user, classify assistant responses, and extract answers from free-text responses. LLM-based components are likely to fail, and we performed an inspection of 200 simulated \shardedt conversations to understand the level of simulation error and the potential effect on estimating the performance of the assistant LLMs due to the error.

For each inspected conversation, we annotated user turns, assistant turns, and the overall conversation with five specific elements.

For user utterances, we annotated whether the utterance revealed exactly the information from one shard in the sharded instruction (\texttt{Shard Fully Revealed}). Specifically, we flagged turns that revealed more than one shard, and turns that revealed a shard only partially. We also annotated each user's turn for whether it is appropriately contextualized in the conversation (\texttt{Shard Contextualized}). For example, if the previous assistant's turn asked a binary clarification question (yes/no), then proper contextualization would require a Yes/No response to directly address the assistant's response.

For assistant utterances, we annotated whether the classified strategy was accurate (\texttt{Strategy Accuracy}). For example, if the response is labeled as a clarification, we confirm if it poses a clarification question to the user. When assistant utterances were labeled as answer attempts, we further labeled whether the answer extraction step was successful (\texttt{Extraction Success}).

Upon completing the inspection of each user and assistance utterance, we assigned a global label to the entire conversation on whether or not the errors that occurred during simulation (if any) affected the overall validity of the simulation. If not, the simulation was marked as successful (\texttt{Overall Success}).

We inspected conversations for four tasks: Actions, Code, Math and Database. The other two (Summary and Data-to-text) are refining tasks that require an answer attempt at each turn, and do not rely on an LLM-based user simulator. As such, they have limited scope for simulation error.

Table~\ref{tab:inspection} summarizes the results of the inspection annotation. Overall, the simulation environment is highly reliable, with roughly 98\% of inspected conversations labeled as successful. Some errors occur in each component. With user simulation, a single shard is fully revealed around 96\% of the time, and properly contextualized 98\% of the time. The processing of assistant responses also leads to errors: the turn strategy classification is only 95\% accurate, and extraction of answer attempts has an accuracy of 97\%.

Utterance-level errors did not always affect the validity of the overall simulation. In some cases, we observed that the user simulator would correct an error in an early turn, subsequently in the conversation, or that an error in answer extraction on the wrong answer attempt would occur at a turn, but the extraction would be successful later on.
In summary, we empirically find that the simulation environment is largely accurate: though some errors occur, large drops of performance in the \shardedt setting (beyond 2\%) are not due to errors caused by the simulator.

\section{Concrete Example of Loss in Aptitude vs. Reliability} \label{app:concrete_a_vs_r}

Let's imagine we are provided with ten instructions ($N=10$), each \fullt and \shardedt. We run simulations with an LLM, simulating 10 conversations per instruction and setting ($M=10$). Let's assume the LLM achieves an averaged performance ($\overline{P}$) of 90\% in the \fullt, and 60\% in the \shardedt setting.

Finally, let's assume that the \fullt performance is achieved by having perfect performance (\textit{i.e.}, success in 10/10 randomly sampled runs) on 9 instructions, and failing on all the sampled simulations of the last, tenth instruction. In other words:

$$
S^{\textsc{Full}}_{ij} = 
\begin{cases}
100, & \text{if } i \in \{1, \dots, 9\} \\
0,   & \text{if } i = 10
\end{cases},
$$

where $S^{\textsc{Full}}_{ij}$ represents the score for $i$-th instruction at $j$-th simulation run.
The aptitude ($A$) and unreliability ($U$) of the LLM for the \fullt setting is $A=90$\% and $U=0$\% (\textit{i.e.}, for each instruction, the 10th and 90th percentile scores are equal).

\begin{wrapfigure}{r}{0.4\linewidth}
    \vspace{-2em}
    \centering
    \includegraphics[width=\linewidth]{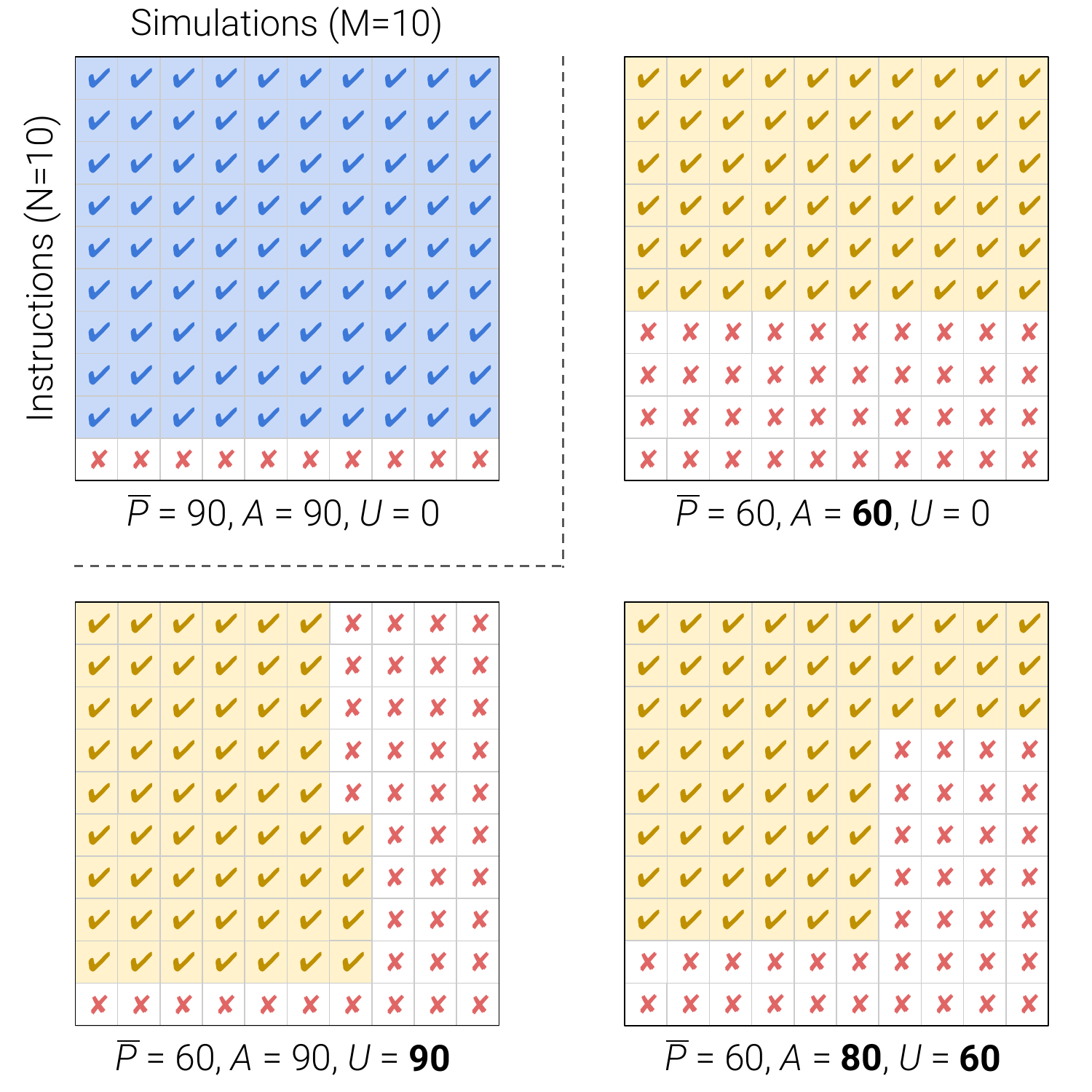}
    \caption{Illustrations for different situations. Green and red fills in each grid indicate sample-level score (\textit{e.g.,} pass / exact match). Compared to \fullt (top left), three situations in \shardedt achieve the same $\overline{P}=60$ while varying in aptitude $A$ and unreliability $U$.}
    \label{fig:metric_illustration}
    \vspace{-7em}
\end{wrapfigure}

Let's now consider three conditions for the \shardedt setting that all achieve an averaged performance of $\overline{P} = 60$\%. We illustrate the conditions in Figure~\ref{fig:metric_illustration}.

\paragraph{Situation 1: Drop in Aptitude.}

The LLM achieves perfect performance on six of the ten instructions:

$$
S^{\textsc{Sharded}}_{ij} = 
\begin{cases}
100, & \text{if } i \in \{1, \dots, 6\} \\
0,   & \text{if } i \in \{7, \dots, 10\}
\end{cases}.
$$

In situation 1, $\overline{P}=60$\%, $A=60$\%, and $U=0$\%. The degradation in performance is entirely explained by a decrease in aptitude, while the reliability remains the same.

\paragraph{Situation 2: Drop in Reliability.}

The LLM achieves mixed performance (6-7 perfect scores per instruction) on nine of the 10 instructions:

$$
S^{\textsc{Sharded}}_{ij} = 
\begin{cases}
100, & \text{if } 1 \leq i \leq 3, 1 \leq j \leq 6 \\
100, & \text{if } 4 \leq i \leq 9, 1 \leq j \leq 7 \\
0,   & \text{otherwise} 
\end{cases}.
$$

In situation 2, $\overline{P} = 60$\%, with an aptitude of $A=90$\%, and a unreliability of $U=90$\%. The degradation in performance is entirely explained by a large drop in reliability, while sharded and fully-specified aptitude are equal.

Situations 1 and 2 illustrate extreme scenarios where the average drop in performance is entirely explained by a drop in aptitude or reliability, but in practice a combination is more likely to occur, as in situation 3.

\paragraph{Situation 3: Combined drop in Aptitude and Reliability.}

The LLM achieves perfect performance on three instructions, and mixed performance (6 perfect scores per instruction) on five of the 10 instructions:

$$
S^{\textsc{Sharded}}_{ij} = 
\begin{cases}
100, & \text{if } 1 \leq i \leq 3 \\
100, & \text{if } 4 \leq i \leq 8, 1 \leq j \leq 6 \\
0,   & \text{otherwise} 
\end{cases}.
$$

In situation 3, $\overline{P} = 60$\%, with an aptitude of $A=80$\%, and a unreliability of $U=60$\%. Note that situation 3 leads to a larger increase in unreliability (from 0\% to 60\%) than a decrease in aptitude (from 90\% to 80\%) when compared to fully-specific simulations. This corresponds in practice to our observation: drops in performance are explained by small drops in aptitude and large drops in reliability.

Finally, we note that though this concrete example we provide uses binary scores (0 and 100) for simulated conversation outcomes, aptitude (A) and unreliability (U) can equally be applied to continuous metrics (such as BLEU).

\section{Qualitative Analyses of Simulation Logs} \label{app:qualitative}

In the following subsections, we report qualitative analyses on the corpus of simulations from the main experiment (Section~\ref{sec:main_results}). The purpose of the analyses is to discern root causes in model behavior that lead to performance degradation. We identify four behaviors below and provide the analysis for each item in the rest of the section: 

\begin{enumerate}
    \item LLMs attempt to answer the entire problem prematurely.
    \item LLMs overly rely on previous (incorrect) answer attempts, leading to lengthier ``bloated'' answers.
    \item LLMs overly adjust their answers based on the last conversation turn, materialized by a pronounced forgetting of middle-turns.
    \item LLMs produce answers that are overly verbose, which likely introduce problem assumptions that detract attention from user-utterances.
\end{enumerate}

\subsection{Premature Answer Attempts} \label{app:premature_answer_attempt}

\begin{wraptable}{l}{0.4\linewidth}
    \vspace{-1em}
    \resizebox{1.0\linewidth}{!}{%
    \setlength{\tabcolsep}{4pt}
    \begin{tabular}{p{2cm}ccccc}

    & \multicolumn{5}{c}{Conversation Progress At First Answer Attempt} \\
    \cmidrule(){2-6}
    Model & 0-20\% & 20-40\% & 40-60\% & 60-80\% & 80-100\% \\

    \cmidrule(lr){1-1} \cmidrule(){2-6}

    \ccol{F5F5F5} \textit{First answer attempt is ...} & \ccol{F5F5F5} \raisebox{-1ex}{earliest}
    & \ccol{F5F5F5} \raisebox{-1ex}{early}
    & \ccol{F5F5F5} \raisebox{-1ex}{midway}
    & \ccol{F5F5F5} \raisebox{-1ex}{late}
    & \ccol{F5F5F5} \raisebox{-1ex}{latest} \\
    \cmidrule(lr){1-1} \cmidrule(){2-6}
    
\llama~3.1-8B & \ccol{FCFDFE}16.1 & \ccol{EDF2FB}24.0 & \ccol{D7E3F7}35.3 & \ccol{CEDDF5}39.6 & \ccol{CEDDF5}39.7 \\
\aitwo~OLMo2 & \ccol{F9FBFD}17.6 & \ccol{DCE6F8}32.7 & \ccol{D2DFF6}37.7 & \ccol{BFD2F2}47.3 & \ccol{E8EFFA}26.4 \\
\claude~3-Haiku & \ccol{E7EEFA}27.1 & \ccol{D6E2F6}35.6 & \ccol{BFD2F2}47.4 & \ccol{A8C2ED}58.9 & \ccol{92B3E9}70.3 \\
\openai~4o-mini & \ccol{E1EAF9}30.2 & \ccol{CFDDF5}39.2 & \ccol{BDD1F1}48.4 & \ccol{AAC3EE}58.2 & \ccol{A7C1ED}59.9 \\
\llama~3.3-70B & \ccol{DBE5F7}33.3 & \ccol{CDDCF5}40.1 & \ccol{B8CDF0}51.2 & \ccol{A6C1ED}60.0 & \ccol{94B4E9}69.3 \\
\microsoft~Phi-4  & \ccol{EAF0FA}25.7 & \ccol{DBE6F7}33.1 & \ccol{C0D3F2}47.0 & \ccol{B4CAF0}53.0 & \ccol{AAC4EE}57.9 \\
\cohere~CMD-A & \ccol{D1DFF5}38.0 & \ccol{C8D8F4}42.9 & \ccol{ADC5EE}56.5 & \ccol{9CB9EB}65.5 & \ccol{8CAEE8}73.5 \\
\llama~4-Scout & \ccol{CEDDF5}39.8 & \ccol{D4E1F6}36.8 & \ccol{B8CDF0}51.0 & \ccol{AAC4EE}57.9 & \ccol{9DBAEB}64.8 \\
\openai~o3 & \ccol{F3F6FC}21.0 & \ccol{D2DFF6}37.9 & \ccol{B6CCF0}51.9 & \ccol{A9C3ED}58.4 & \ccol{97B6EA}68.0 \\
\claude~3.7-Sonnet & \ccol{E3EBF9}29.2 & \ccol{D6E2F6}35.6 & \ccol{AFC7EF}55.3 & \ccol{96B6EA}68.0 & \ccol{8FB1E8}71.6 \\
\deepseek~R1 & \ccol{CEDDF5}39.5 & \ccol{C7D8F3}43.1 & \ccol{B3CAEF}53.5 & \ccol{9AB8EA}66.4 & \ccol{B9CEF1}50.2 \\
\openai~4o & \ccol{D5E2F6}36.0 & \ccol{CBDAF4}41.4 & \ccol{AEC6EE}56.2 & \ccol{9BB9EB}65.6 & \ccol{7FA5E5}90.4 \\
\gemini~2.5-Flash & \ccol{CFDEF5}39.0 & \ccol{BDD0F1}48.6 & \ccol{A6C0ED}60.2 & \ccol{91B2E9}70.8 & \ccol{8AADE7}74.6 \\
\openai~4.1 & \ccol{D9E5F7}33.9 & \ccol{B5CBF0}52.7 & \ccol{A5C0ED}60.6 & \ccol{95B4E9}69.0 & \ccol{82A7E6}78.6 \\
\gemini~2.5-Pro & \ccol{CBDBF4}41.1 & \ccol{C2D4F2}45.7 & \ccol{B3CAEF}53.5 & \ccol{9DBAEB}64.6 & \ccol{9FBCEB}63.8 \\

\cmidrule(lr){1-1} \cmidrule(){2-6}

Average & \ccol{DFE9F8}30.9 & \ccol{CCDBF4}40.5 & \ccol{B6CCF0}51.7 & \ccol{A5C0ED}60.4 & \ccol{9EBBEB}64.4 \\

    \bottomrule
    \end{tabular}
    }
    \caption{Averaged performance ($\overline{P}$) breakdown, based on how early in the conversation the LLM makes its first answer attempt. Analysis conducted on simulations of two tasks: Code and Math.}
    \label{tab:first_answer_attempt_earliness}

    \vspace{-3em}
\end{wraptable}

During \shardedt simulation, responses are classified according to a seven-class conversation strategy categorization. In particular, each assistant response is tagged as being a formal \textit{answer attempt} or not (as answer attempts require further processing: extraction and evaluation by the task-specific evaluator).

On the onset of conversation, LLMs have the least amount of information (highest level of underspecification), and are least likely to formulate correct answer attempts. Proposing a solution early might therefore plant certain incorrect elements in it, which wrongly influences the interaction later in the conversation.

To evaluate this hypothesis, we bin all simulated conversations from our experiments based on how early in the conversation the first answer attempt is generated by the LLM. Specifically, we create five bins: \texttt{0-20\%} if the first answer attempt occurs within the first 20\% turns of the conversation, and \texttt{20-40\%}, \texttt{40-60\%}, \texttt{60-80\%}, and \texttt{80-100\%} if it occurs in later turns of the conversation.
Of the six tasks included in our experiments, only two (Math and Code) observed a significant range in LLM behavior for answer attempt timing. For the other four tasks, models attempt an answer from the first turn in most of the time, rendering analysis on this parameter impossible.

Analysis results for the two remaining tasks are presented in Table~\ref{tab:first_answer_attempt_earliness}. We observe that for every single model, conversations with a later first answer attempt lead to higher averaged performance. Across all models, conversations with the first attempt being made in the first 20\% of conversations achieve a score of 30.9, less than half of the 64.4 when the LLM waits for the last 20\% of the conversation to make an answer attempt.

In other words, we find that premature answer attempts detract LLM performance. Conversations where the model clarifies user instructions or discusses the problem at a high-level before moving to generating complete answer attempts lead to higher performance. We hypothesize that this is due to the model making incorrect assumptions in premature solutions, which conflict with subsequent user instructions in later turns.

\subsection{Answer Bloat in Multi-Turn Conversation} \label{app:answer_bloat}

\begin{figure}[ht]
    \centering
    \includegraphics[width=0.98\linewidth]{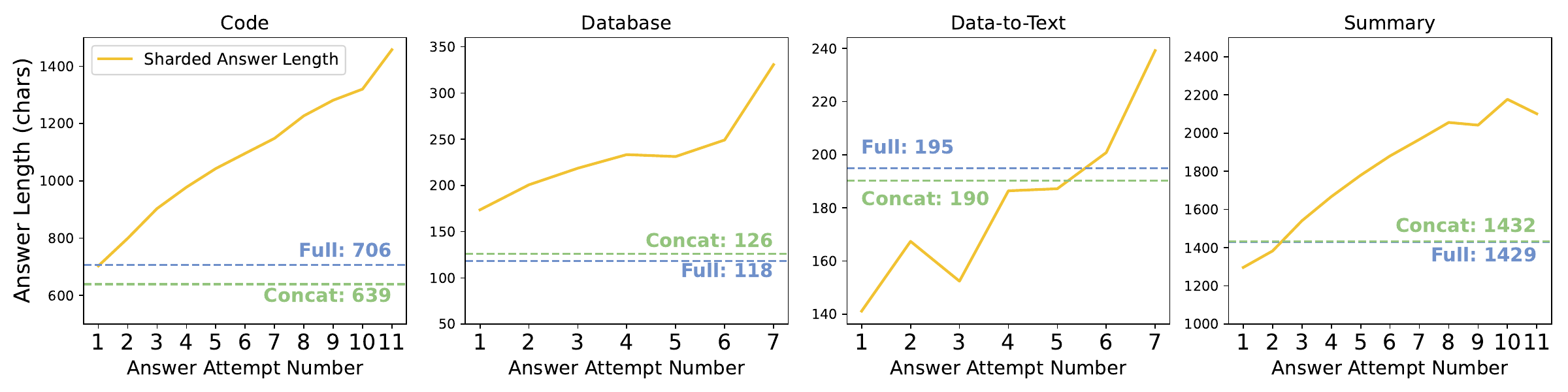}
    \caption{Average length (in number of characters) of answer attempts across four tasks (Code, Database, Data-to-text, and Summary) in \shardedt conversations. Answer attempts in the \fullt and \concatt settings tend to be shorter on average than those from \shardedt setting. \shardedt answer attempts increase in length as the LLMs make more answer attempts.}
    \label{fig:answer_attempt_length}
\end{figure}

In multi-turn conversation simulations, the LLM might make multiple answer attempts, with each subsequent attempt being potentially based on previous attempts. In contrast, single-turn conversations constrain conversation dynamics, with the LLM making a single, first-and-final answer attempt.

To understand multi-turn conversation dynamics, we calculate the average length of answer attempts in each simulation type. For the \shardedt setting, we calculate average length for each attempt within a simulation (\textit{i.e.}, average length of the first attempt, second attempt, third attempt, etc.). We note for readers here that the analysis is conducted on extracted answer attempts (output of the Answer Extractor module in Figure~\ref{fig:simulator}) rather than the entire assistant responses. The extracted answer more accurately measures dynamics in answer attempts (i.e., generated SQL query, or Python function) rather than the entire responses, which might contain varying amounts of unrelated content.

Results of the analysis are plotted in Figure~\ref{fig:answer_attempt_length}. Across the four tasks, we find that answer lengths in the \fullt and \concatt settings tend to be similar, typically within 2-10\% of each other. On three of the analyzed tasks (Code, Database, Summary), the first answer attempt in the \shardedt setting has a similar length to \fullt and \concatt counterparts, yet for each subsequent answer attempt, we observe an increase in average answer length. The effect is such that the final answer attempts in \shardedt conversations (right portion of the four plots) tend to be 20-300\% longer than the solutions generated in the \fullt and \concatt settings. We name this observation the \textit{answer bloat effect}: as a multi-turn conversation progresses, the LLM generates incorrect answer attempts, making assumptions about portions of the instruction that remain unspecified. As the user reveals additional information in succeeding turns, the LLM does not successfully invalidate its prior assumptions and overly relies on its previous attempts. Answer bloat in multi-turn, underspecified conversation leads to longer solutions compared to single-turn equivalents.

We perform an additional analysis, focusing only on the Code and Database tasks and filtering to simulations where the LLM reaches an entirely correct solution (score of 100.0). For Code task, correct programs obtained from \shardedt setting are on average 850 characters long, which is 27\% more characters than the correct solutions generated in the \fullt setting (668 characters on average).
For Database, correct SQL queries in the \shardedt setting are on average 129 characters, 14\% more characters than those from the \fullt setting (113 characters).
In summary, LLMs are less likely to reach a correct solution in multi-turn settings (lower $\overline{P}$), and when they do, the final solutions they reach are longer (bloated), hinting that the solutions are qualitatively worse.

\subsection{Over-adjust based on Last Turn of Conversation} \label{app:summary_last_turn}

Because the summary task requires the assistant to attribute its summary back to documents through citation, the task offers a unique opportunity to analyze what turns of information LLMs pay attention to as the multi-turn conversation progresses. As a reminder, the summary task involves a user introducing new documents at each turn. The focus of our analysis is therefore to understand whether document introduction order (across turns) affects the likelihood of the LLM citing a document.

\begin{wrapfigure}{r}{0.35\linewidth}
    \centering
    \includegraphics[width=\linewidth]{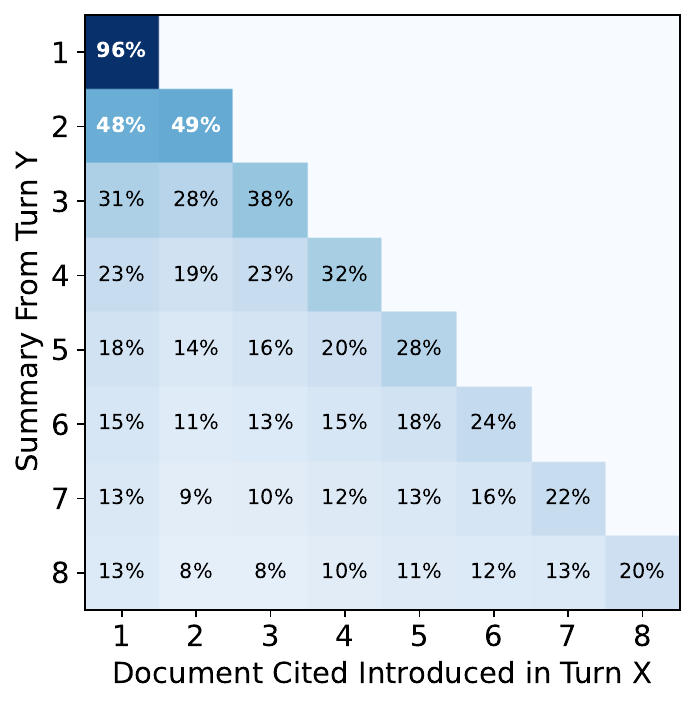}
    \caption{Analysis of citation patterns in summaries generated by LLMs with the \shardedt simulation. At each turn, the LLM generates an updated summary (y-axis), which includes citations from the documents that have been revealed up to this turn. Percentages in a row do not add up to 100\% due to citation hallucinations that occur for some models.}
    \label{fig:summary_citation_analysis}
    \vspace{-5em}
\end{wrapfigure}

In Figure~\ref{fig:summary_citation_analysis}, we plot the the results of our analysis. Each row corresponds to the analysis of summaries generated at a given turn in the sharded simulation. At turn 1 (top row), 96\% of the cited documents were introduced in the first turn. The missing 4\% correspond to hallucinated citation to documents that were not introduced, and explains why none of the rows' distribution sum to 100\%. At turn two (second row from the top), summaries include citation in roughly equal proportion for turn-1 and turn-2 documents (i.e., 48\% and 49\%). 
We interpret this to mean that in 2-turn conversations, LLMs pay roughly equal attention to documents in either turn.
Analysis of summaries generated in turns 3-8 of sharded simulations reveal an imbalance in the documents the LLM cites to. In eighth-turn summaries, 20\% of citations are to documents introduced in turn 8, compared to 8\% from turn 2 and 3 (150\% difference). At a high-level, as the conversation progresses, LLMs are most likely to cite either documents in the first or last turns, and less likely to cite documents introduced in intermediary (middle) turns. This finding mirrors findings of a \textit{loss-in-the-middle} phenomena of LLMs paying more attention to documents at the start or end of their provided context, at the cost of middle-context content \cite{huang2023embrace,liu2024lost,laban2024summary}. In short, we observe that the lost-in-the-middle phenomena occurs not only in single-turn long-context settings, but also in multi-turn conversation. We name this phenomenon \textit{loss-in-middle-turns}.

We note that the analysis presented in Figure~\ref{fig:summary_citation_analysis} averages numbers across the 15 LLMs included in our main experiment. Even though we observe some loss-in-middle-turns in all models, the magnitude of the effect varies across models, typically with more performant models having a more muted effect, showing they have better capabilities of handling attribution across multiple turns of context. We do not include model-specific analyses in this work and leave it for future work.

\subsection{Overly-verbose Assistant Responses} \label{app:assistant_verbosity}

\begin{wraptable}{l}{0.5\textwidth}
    \vspace{-1em}
    \resizebox{1.0\linewidth}{!}{%
    \setlength{\tabcolsep}{4pt}
    \begin{tabular}{p{2cm}ccccc}
    & \multicolumn{5}{c}{Relative Assistant Verbosity} \\
    \cmidrule(){2-6}
    Task & 0-20\% & 20-40\% & 40-60\% & 60-80\% & 80-100\% \\
    \cmidrule(lr){1-1} \cmidrule(){2-6}
    \textit{Assistants responses are ...} &
    \ccol{F5F5F5} \raisebox{-0.7ex}{shortest} &
    \ccol{F5F5F5} \raisebox{-0.7ex}{short} &
    \ccol{F5F5F5} \raisebox{-0.7ex}{median} &
    \ccol{F5F5F5} \raisebox{-0.7ex}{long} &
    \ccol{F5F5F5} \raisebox{-0.7ex}{longest} \\
    \cmidrule(lr){1-1} \cmidrule(){2-6}
    Code & \ccol{84A9E6}55.3 & \ccol{9DBAEB}52.3 & \ccol{B7CDF0}48.9 & \ccol{C7D8F3}46.9 & \ccol{EBF1FB}42.5 \\
    Math & \ccol{91B2E9}62.9 & \ccol{80A6E5}64.0 & \ccol{9DBBEB}62.1 & \ccol{B0C7EF}60.9 & \ccol{FCFDFE}56.1 \\
    Database & \ccol{81A7E5}43.8 & \ccol{A6C1ED}40.0 & \ccol{C1D3F2}37.3 & \ccol{DEE8F8}34.3 & \ccol{FCFDFE}31.3 \\
    Actions & \ccol{F9FBFD}41.5 & \ccol{ABC4EE}49.6 & \ccol{7FA5E5}54.2 & \ccol{83A8E6}53.6 & \ccol{9EBBEB}50.8 \\
    Data-to-Text & \ccol{81A6E5}25.0 & \ccol{A3BEEC}24.3 & \ccol{B1C8EF}24.0 & \ccol{DFE8F8}23.1 & \ccol{FEFDFD}21.8 \\
    Summary & \ccol{7FA5E5}15.4 & \ccol{88ABE7}14.7 & \ccol{A5C0ED}13.5 & \ccol{CCDBF4}12.0 & \ccol{F7F9FD}10.3 \\
    \cmidrule(lr){1-1} \cmidrule(){2-6}
    Average & \ccol{96B5EA}40.7 & \ccol{94B4E9}40.8 & \ccol{A2BEEC}40.1 & \ccol{BED1F2}38.6 & \ccol{F4F7FC}35.6 \\
    \bottomrule
    \end{tabular}
    }
    \caption{Averaged performance ($\overline{P}$) of LLMs on the six experimental tasks, arranged based on model relative verbosity (length of response). Performance degrades when models generate longer responses on five of the six tasks.}
    \label{tab:relative_verbosity}
    \vspace{-1em}
\end{wraptable}

When simulating multiple conversations based on a common instruction, we observe variation in responses, particularly in the length of the response generated by the LLM. To understand how verbosity (length of a response) affects model performance, we perform a verbosity analysis.

One difficulty with assessing verbosity is that different tasks and instructions might require different levels of verbosity. For example, generating a Python function likely requires a longer than generating an SQL query. In order to regularize for task-specific variations, we assign a \textit{verbosity tag} calculated for each (LLM, instruction) tuple. For each simulated sharded conversation involving an LLM on an instruction, we calculate the average length of the per-turn response (number of total characters in assistant responses divided by number of turns). We then bin conversations into quintiles according to this metric. More specifically, since we simulated $N=10$ conversations for each (model, instruction) pair, we assign 2 simulations per quintile, which we name: shortest, short, median, long, and longest. We then calculate averaged performance ($\overline{P}$) on the six experimental tasks, arranged based on this verbosity tag. Results are summarized in Table~\ref{tab:relative_verbosity}.

On five of the six tasks, performance is 10-50\% higher in simulated conversations with shortest response length, compared to conversations with longest response length. As assistant responses get longer (left to right in the Table), performances gradually drop. The Actions task is the only task where such an effect is not observed, and where shortest response length from the assistant is detrimental.

Predominantly however, models achieve higher performance when they generate shorter responses. We hypothesize that deterioration due to over-verbosity is due to longer responses typically containing more assumptions or hypotheses from the assistant, which can lead to confusion in following turns. On the other hand, short turns tend to be focused (e.g,  a single clarification question), and keep the conversation on track.

Deterioration due to over-verbosity is note-worthy, as besides deteriorating underlying model performance, longer responses also take longer for users to read, which is undesirable. The finding therefore indicates that longer LLM responses are bad both for models and end-users.

\section{Assistant Response Categorization} \label{app:answer_categorization}
We categorize each assistant response into one of the seven categories to capture the answer attempt and evaluate if that is the case, as well as to understand the model behavior tendency.
\citet{herlihy2024overcoming} defined seven turn categories for LLM responses and classified them using LLM, uncovering that GPT-4 prefers answering directly even when the query is underspecified.
Motivated by this study, we similarly define seven response categories which we list in Table~\ref{tab:answer_categorization}, together with example responses.
Key differences are discussion and answer attempt; we observed many responses containing large body of text formulating the question in our preliminary experiments, which led to redefining ``Miscellaneous'' from \citep{herlihy2024overcoming} into ``Discussion'' in our experiment.
``Direct Response'' in \citep{herlihy2024overcoming} corresponds to our ``Answer Attempt.''

\begin{table}[t]
\centering
\begin{tabular}{@{}lp{0.4\textwidth}p{0.4\textwidth}@{}}
Name            & Description  & Example \\ \midrule
Answer attempt  & The response contains a complete answer attempt to the question that can be extracted verbatim.  & The dog is 50 meters away from the house.\\\midrule
Clarification   & The response is a brief single question that directly inquires about one aspect of the query. & To calculate the distance, I need to know how long the dog ran. Could you provide more information about that? \\\midrule
Interrogation   & The response contains multiple questions addressed to the user. & I cannot answer the question without knowing (1) speed, (2) duration, and (3) starting position. Please tell me about these points and I can calculate the distance!  \\\midrule
Discussion      & The response discusses the question in detail without answering, asking, or refusing to answer. & The question is trying to measure the distance between the dog and the house. We can calculate based on this equation: [Equation]. [$\ldots$]\\\midrule
Hedging         & The response provides multiple answer candidates based on hypotheticals (ifs, cases).   & 1. If the dog was originally in the house, it would be 50 meters away now.\newline 2. If the dog was at the park, it would be 100 meters away from the house now.        \\\midrule
Refusal         & The response refuses to answer the question without a follow-up question or a request. & I can't answer your question because I don't have sufficient information.\\\midrule
Missing         & The response is empty. & [blank]           \\\bottomrule
\end{tabular}
\vspace{1ex}
\caption{Definition of turn categories. We include the description in the prompt to categorize assistant responses.}
\label{tab:answer_categorization}
\end{table}

\section{Model Access}\label{app:model_details}

We accessed models that were used in the experiments from various vendors. 
The short form names we used throughout the paper, the corresponding versions, and the providers are summarized in Table~\ref{tab:model_list}.
Except for the exploration with various temperatures (Section~\ref{sec:implications_llms}), we set the temperature to $T=1.0$ and used the default values for the rest of configurable hyperparameters. We set the maximum response length to 1,000 tokens for all models, and did not observe models exceeding this limit frequently when generating responses. For thinking models (o3, Deepseek-R1), we increased the limit to 10,000 tokens to account for the additional test-time compute (thinking tokens).

\begin{table}[tbhp]
\centering
\begin{tabular}{@{}llll@{}}
Short Form & Name                         & Version & Access Provider \\ \midrule
\openai~4o & GPT-4o               & \texttt{gpt-4o-2024-11-20}        & OpenAI / Microsoft API\\
\openai~4o-mini & GPT-4o-mini          & \texttt{gpt-4o-mini-2024-07-18}  & OpenAI API \\
\openai~4.1 & GPT-4.1              & \texttt{gpt-4.1-2025-04-14}        & OpenAI / Microsoft API\\
\openai~o3 & o3                   & \texttt{o3-2025-04-16}        & OpenAI / Microsoft API \\\midrule
\claude~3-Haiku & Claude 3 Haiku    & \texttt{claude-3-haiku-20240307}    &  Amazon Bedrock      \\
\claude~3.7-Sonnet & Claude 3.7 Sonnet & \texttt{claude-3-7-sonnet-20250219} &  Amazon Bedrock      \\\midrule
\gemini~2.5-Flash     & Gemini 2.5 Flash     & \texttt{gemini-2.5-flash-preview-04-17}        &  Gemini API      \\
\gemini~2.5-Pro     & Gemini 2.5 Pro       & \texttt{gemini-2.5-pro-preview-03-25} &  Gemini API      \\\midrule
\llama~3.1-8B & Llama-3.1-8B-Instruct  & N/A     & Local Ollama \\
\llama~3.3-70B & Llama-3.3-70B-Instruct & N/A     &  Amazon Bedrock      \\
\llama~4-Scout & Llama-4-Scout-17B-16E  & N/A     &  Together AI      \\\midrule
\cohere~CMD-A & Command-A & \texttt{command-a-03-2025}     & Cohere API \\
\deepseek~R1 & Deepseek-R1        & N/A     &  Amazon Bedrock      \\
\aitwo~OLMo2 & OLMo2-13B         & N/A     & Local Ollama \\
\microsoft~Phi-4 & Phi-4   & N/A     & Local Ollama \\
\bottomrule
\end{tabular}
\vspace{1ex}

\caption{Specific model versions used as part of our experiments. For each model, we define the exact \texttt{Version} of the model accessed (for models that have versioning) and the \texttt{Access Provider} to facilitate result reproducibility.}
\label{tab:model_list}
\end{table}

\section{Task-specific Implementation details} \label{app:task_implementation}

We provide task implementation details. For each task, we specify: (1) the selection of original single-turn fully-specified instruction, (2) the evaluation metric that was repurposed from the original dataset, (3) and what the initial system messages consists of (if any). 

\subsection{\texorpdfstring{\python Code}{Code}}

The Code instructions are sourced from a combination of HumanEval~\cite{chen2021evaluating}, a dataset of 164 basic Python programming problems given the function header and the docstring that specifies the problem, and LiveCodeBench~\cite{jain2024livecodebench}, an evolving dataset of Python algorithmic challenges.
In particular, we source from the ``call-based'' problem subset in LiveCodeBench v5, with the difficulty of either ``Easy'' and ``Medium'', to align the solution formats between the two sources.

We first sharded all HumanEval problems following the protocol mentioned in Appendix~\ref{app:sharding_process}, obtaining 45 high quality sets of shards that meet the criteria. The rest of the dataset were discarded because of being simplistic, leaving little room to construct sufficient number of shards for a problem.
Subsequently, we shuffled and sharded the aforementioned subset from LiveCodeBench until obtaining 100 valid sharded instructions.

We follow the original prompts used by the benchmark authors as much as possible for the single-turn (\fullt and \concatt) evaluation. 
Specifically, \fullt prompt from HumanEval includes the function header and the docstring provided as \texttt{prompt} in HumanEval dataset, and \fullt \& \concatt from LiveCodeBench includes \texttt{starter\_code} consisting of the function signature.

Both HumanEval- and LiveCodeBench-derived problems come with test cases which we use to compute the functional accuracy of the answer attempt by the LLMs.
We re-use the evaluation codebase maintained by~\citet{jain2024livecodebench}, which (1) wraps the candidate function in a test module, (2) execute given the inputs, and (3) checks the equivalence of the output from the expected output, with a default timeout set to prevent the evaluator from getting trapped during evaluation (\textit{e.g.}, brute-force implementation may not pass under the set time budget).
In case when multiple code blocks are present in a response, the answer extraction module selects the last function definition in the last markdown code block.

\subsection{\texorpdfstring{\database Database}{Database}}

The Database instructions are sourced from the validation portion of the Spider dataset \cite{yu2018spider}. We note that though a more recent version of Spider has been released (Spider 2.0 \cite{lei2024spider}), the instructions in the second iteration are more advanced and represent less typical database use, and we select instructions from the more realistic Spider 1.0.

The authors of Spider categorized queries into four levels of difficulty (EASY, MEDIUM, HARD, XHARD), based on the syntax complexity of a reference SQL query. We filtered out queries of EASY complexity, as they tended to yield fewer than three shards when processed. The rest of the 433 natural language queries in Spider were gradually sharded until reaching a total of 107 valid sharded instructions.

Each original instruction in Spider supplies a database schema, represented in SQL as a series of table schema (i.e., each define a series of columns including name, type, and optional index). We include the database schema as part of the system message (i.e., prior to the first turn of conversation), and informing the LLM that users will provide natural-language queries that must be answered using a database with the provided schema. 

Each original instruction in Spider is paired with a reference SQL solution. We follow \citet{zhong2020semantic} for the evaluation methodology. For a given original instruction, the candidate and reference SQL queries are executed on a fixed set of databases, and exact match of the results on all databases is required to mark the candidate as successful (Score = 100). If a discrepancy is observed on any test database, the candidate is incorrect (Score = 0). One limitation of SQL execution is that false positives can occur: two queries can return the same output on a given database, even when they are not semantically equivalent. \citet{zhong2020semantic} found that by evaluating on an  increased number of databases, false positives become negligible. Finally, any invalid candidate that does not successfully execute (e.g., syntax error) is considered incorrect (Score = 0).

\subsection{\texorpdfstring{\apis Actions}{Actions}}

The Actions instructions are sourced from the released test portion of the Berkeley Function Calling Leaderboard V3 (BFCL) \cite{2024bfcl}. BFCL V3 consists of three sub-genre of instructions: (1) Parallel, (2) Multiple, and (3) Multiple-Parallel. Initial experimentation with the sub-genres identified Parallel as the most suited for sharding, as Parallel instructions specify multiple subtasks that should be used and combined into a single action that accomplishes the entirety of the instruction. We shuffled all the BFCL V3 Parallel instructions, and sharded gradually until we obtained 105 valid sharded instructions.

We note that though a more recent iteration of BFCL includes multi-turn instructions, it differs from sharding experiments as it does not involve underspecification, with each turn having an independent intermediate solution (which we call episodic multi-turn conversations). Our implementation in comparison shards original instructions allowing us to simulate multi-turn underspecified conversations for this task setting. The Background section (Section~\ref{sec:background}) discusses the relationship between episodic and underspecified multi-turn conversation more in-depth.

Each instruction in BFCL comes with tool set documentation, a JSON object that specifies the set of available actions (APIs) for the assistant to complete user instructions. We include the tool set documentation as part of the system message, along with a message indicating that user queries will require the use of the provided tools to be completed.

Each instruction in BFCL comes with a reference answer, consisting of the API calls that should be called to accomplish the user instruction. The maintainers of BFCL have released an evaluation toolkit that assesses semantic equivalence between a candidate answer and the reference answer. We leverage the official evaluation toolkit, assigning a score of S=100 for candidate answers that are considered semantically equivalent to the reference answer, and a score of S=0 otherwise. When the evaluation toolkit is not able to parse a candidate answer (e.g., a syntax error), the candidate is considered incorrect (S=0).

\subsection{\texorpdfstring{\tmath Math}{Math}}

The Math instructions are sourced from the ``main'' portion of the GSM8K dataset \cite{cobbe2021training}. We did not perform a filter on the original 8,700 instructions. We shuffled the instructions and sharded incrementally until we obtained 103 valid sharded instructions. Each GSM8K is paired with a numerical reference answer. We used the official toolkit released alongside GSM8K to standardize numerical answers (i.e., strip formatting, etc.). Standardized candidate numerical answers can then be compared through exact match to the reference answer. If the toolkit detects a match, the candidate answer is considered correct (Score=100), and incorrect otherwise (Score = 0). A short, single-sentence system prompt is used to indicate to the assistant that it will be solving mathematical problems.

\subsection{\texorpdfstring{\datattext Data-to-Text}{Data-to-Text}}

The Data-to-Text instructions are based on instructions in the released test set ToTTo dataset \cite{parikh2020totto}. In ToTTo, fully-specified instructions have the following information elements: (1) a HTML-formatted table extracted from a Wikipedia page, (2) a subset of cells in the table that have been \textit{highlighted}, (3) the name of the Wikipedia page that included the Table, (4) the name of the Section in the Wikipedia page that included the Table. Given these elements, the task objective is to generate a caption for the Table specifically focusing on the highlighted cells and considering the available meta-data. Instructions were shuffled and sharded incrementally until we obtained 120 valid sharded instructions.

For each instruction, we generate sharded instructions by assigning different information elements to individual shards. The first shard consists of the initial HTML-formatted table without highlighting. The second shard provides an updated table with the highlighting present, the third shard provides the Wikipedia page name, the fourth shard provides the Wikipedia Section name. Finally, a fifth shard provides a fixed set of 10 randomly-selected example captions from the training set of the ToTTo dataset.

Each instruction in ToTTo is assigned one to three reference captions that were collected by authors of the original dataset. Evaluation on a candidate caption calculates the BLEU score \cite{papineni2002bleu} between the candidate and the set of available references, following the evaluation methodology from the original paper.

The Data-to-Text is a refinement task; at each turn, the model is provided an additional shard of information, and is explicitly told to update its response considering all the information provided so far. As a refinement task, assistant responses at each turn are automatically categorized as answer attempts, and the extracted answer is considered to be the entire response. The system instruction informs the model that its response should consist solely of a table caption, without additional text (such as intro, outro, or politeness wording).

\subsection{\texorpdfstring{\summary Summary}{Summary}}

The Summary instructions are based on samples of the Summary of a Haystack dataset \cite{laban2024summary}. We reuse the entire instructions from Summary of a Haystack to produce 92 sharded instructions. The original instructions each consist of a \textit{haystack} -- 100 documents for a total of 100,000 tokens of content -- and a user query. The goal of the task is to generate a bullet-point-formatted summary of the query-relevant insights that occur in the collection of documents, and use citation to attribute information in each of the bullet points back to the source documents.

The original setting of the Summary of a Haystack purposefully includes a large amount of redundancy (each insight is repeated across at least 6 documents) to evaluate LLMs' ability to thoroughly cite sources. However, we simplify the task for the multi-turn setting, as the 100,000-token haystacks restrict the variety of models we can evaluate. We instead follow subsequent work in selecting smaller Haystacks (``mini-Haystacks'') \cite{belem2024single}. Mini-Haystacks consist of 20 documents and ensure that each reference insight is repeated across three documents. For each instruction, we produce ten shards by randomly assigning two documents per shard. The initial shard further specifies high-level task instruction, by specifying the user query, the expected bullet-point format, with a formatted citation.

Summary of a Haystack relies on an LLM-based metric (Joint Score) to compute the quality of the summary in terms of both the relevance of the candidate bullet points (coverage) and the quality of the generated attribution within the bullet points (citation). The authors note that the metric is recall-based, such that longer summaries are likely to score higher than shorter ones. To account for length bias, the original task instructs models to generate summaries of at most 300 words, which we include in our experiments as well. Specifically, models are instructed in all settings to generate summaries of up to 300 words. We observed that in multi-turn settings, models often \textit{forget} this instruction, leading to non-adherence to the instruction. To avoid penalizing models that correctly remain within the 300-word limit, we truncate summaries that go beyond the limit, removing words in equal proportion from summary bullet points, such that evaluated summaries all respect the 300-word limit. We note that this tendency for LLMs to go beyond is further discussed in Appendix~\ref{app:qualitative}, where we observe that across tasks, model answer attempts get ``bloated'' over turns of conversations. In single-turn settings (full, concat), LLMs largely respect the 300-word length limit.

The summary task is a refinement task. Assistant responses at each turn are automatically categorized as answer attempts, and the entire response is considered to be the extracted answer.

\subsection{\texorpdfstring{\translation Translation}{Translation}}

The Translation instructions were collected from the WMT 2019 task on document-level translation \cite{scherrer2019wmt}. Specifically, we selected 30 documents German-English pairs. Document pairs are aligned at the sentence level (i.e., English and German documents in a pair have the same number of sentences). We truncated the selected pairs to their first ten sentences, and sharded the document instruction such that each shard would introduce exactly two sentences from the document, for a total of five shards. We provided shards in German, and the task consisted in translating into English (i.e., German$\to$English). Hence, Shard 1 introduces the first two German sentences, Shard 2 introduces German sentences 3-4, etc. In the sharded setting, the task requires the LLM to translate the document with all the provided sentences so far. In the full settings, the LLM is provided the entire document (10 sentences) in the first turn. In the concat setting, the LLM is also provided all sentences in the first turn, but separated into the shards (two sentences at a time).

In initial experiments, we experimented with other sharding strategies, including breaking shards at a specific number of words (rather than sentence boundary), and increasing the length of documents (from 10 to 20 sentences), without observing significant differences in results. This led us to adopt the setting we describe: sharding every two sentences, and truncating at 10-sentences.

We evaluated performance with the BLEU metric\cite{papineni2002bleu}, the standard metric for translation tasks, which was used as well in the original WMT 2019 competition.

\section{Example Simulated Conversation} \label{app:example_conv}

\begin{figure}[tb]
    \centering
    \includegraphics[width=\linewidth]{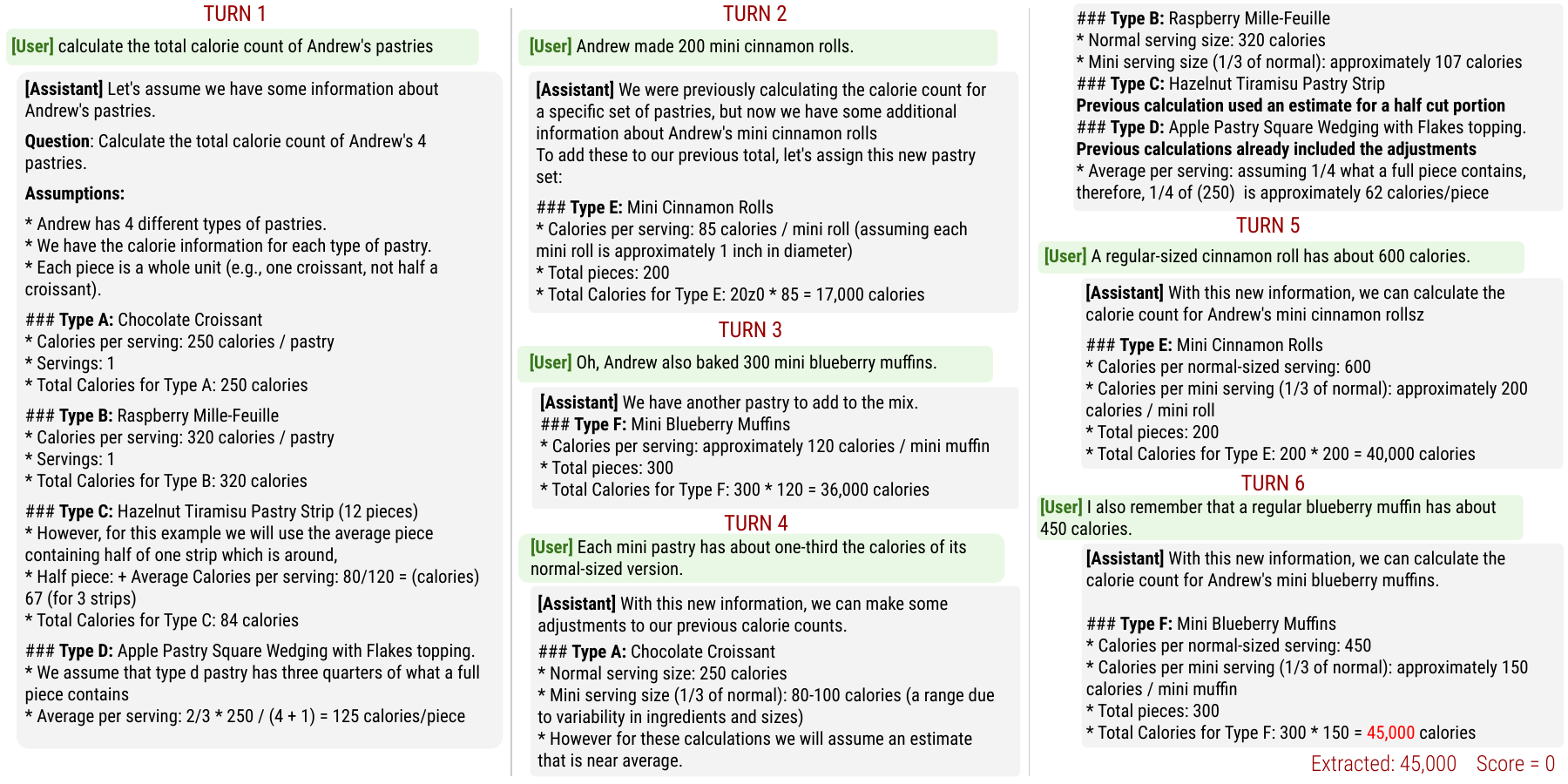}
    \caption{Example simulated multi-turn conversation for the Math task. This conversation simulation was with assistant model Llama3.1-8B-Instruct. The sharded instruction consists of six shards. The correct answer to the instruction is 85,000 calories.}
    \label{fig:example_conv}
\end{figure}

Figure~\ref{fig:example_conv} provides an example conversation that was simulated during our experiments in the sharded setting. The simulation was conducted on the Math task, with a 6-shard instruction, and using the Llama3.1-8B-Instruct as the assistant. This conversation illustrates the following properties described in the rest of the paper: (1) the LLM makes assumptions early in the conversation (in Turn 1, describing four pastries that are irrelevant), (2) although it correctly interprets user-provided information, it also unnecessarily updates the information for assumptions it made (Turn 4), (3) this leads to unnecessary complexity, and the model ultimately forgets that the initial instruction was to calculate total calorie count, and returns only half of the calculation (just for Mini Blueberry Muffin). In short, this conversation illustrates the lost in conversation phenomenon: when the user instruction is underspecified (Turns 1-4), the LLM makes assumptions that detract from the conversation and lead to incorrect or incomplete answers.

\section{Gradual Sharding Implementation} \label{app:gradual_sharding}

To evaluate the effect of instruction granularity on performance degradations, we conducted the \textit{gradual sharding experiment}.

We selected sharded instructions that had exactly eight shards, leading to a total of eight instructions across three tasks (Code, Math, Data-to-Text). We then leveraged an LLM (GPT-4o) to expand each instruction into 7 variants with differing number of shards. The LLM was instructed to \textit{merge} the original sharded instruction into a smaller sharded instruction with two to seven shards. The instruction authorized minor rephrasing to allow for individual shards to be fluent, but encouraged the LLM to remain as close as possible to the original instruction in wording.

As such, each of the original instruction can be paired to: (1) a concat instruction (one-shard), and (2) 7 sharded instructions, ranging from two to eight shards. Applying this method to the 31 instructions yields a total of 248 instructions, with an equal number for the number of shards (from 1 to 8) and on the identical underlying problems.

We ran simulations using the 248 instructions, simulating 10 conversations per instruction and model for two models: GPT-4o and GPT-4o-mini. Findings of the gradual sharding experiment are described in Section~\ref{sec:gradual_sharding}.

\section{Temperature Experiment Implementation} \label{app:temperature}

To evaluate the effect of temperature on aptitude and reliability of LLMs in single- and multi-turn settings, we conducted the following \textit{temperature experiment}.

We selected 10 instructions from each of four tasks: Code, Database, Actions, and Math (for a total of 40). We ran experiments with two models (GPT-4o and GPT-4o-mini). For each instruction and each temperature combination, we conducted simulations for three conversation settings: full, concat, and sharded. For each conversation setting, we varied temperature parameters to three values: 0.0, 0.5, and 1.0. For the full and concat setings, this corresponds to three temperature combinations (as only the assistant temperature can be modified), whereas there are a total of nine combinations for the sharded setting, as both the assistant and user temperature is varied.

We chose to increase the number of simulations to 20 runs per condition (compared to 10 in the main experiment), as the focus of the experiment is to measure variations in model aptitude and reliability, and added simulation runs lead to better percentile estimates used in calculating metrics. This added requirement was not computationally expensive as the temperature experiment involved a limited number of models (2 vs. 15) and instructions (40 vs. 600) in comparison to our main experiment.

Findings of the experiments are described in Section~\ref{sec:implications_llms}.

\section{Recap \& Snowball Experiment Implementation} \label{app:recap_snowball}
We leverage \shardedt conversation logs to simulate \recapt setting, since \recapt only differs from \shardedt in terms of an additional recapitulation turn that gathers all the previous user utterances.
This implementation also allows us to directly compare the effect of the approach against the \shardedt results.
Specifically, for each \shardedt simulation run, we appended the ``recap'' turn and run the simulation one more turn.
Since it requires stacking the past turns every turn, we simulate the entire conversations from scratch for \snowballt simulations.
The prompt concatenates the previous turn user utterances as bullet points, followed by the text for the current turn: \verb|Just to reiterate:\n - [past utterance 1]\n- [past utterance 2]\n\n Also,\n[current utterance]|.
We note that what is accumulated for both \recapt and \snowballt are verbalized utterances from the user simulator, not the original shards themselves.
For both simulation settings, we run $N=10$ simulations on all of the sharded instructions on four tasks (Code, Database, Math, Actions) and report the mean of averaged performance over the tasks, which is shown in Table~\ref{tab:recap_snowball}.

\section{On obtaining deterministic outputs from LLMs} \label{app:randomness}

As we demonstrated in our experimental results, setting the temperatures to zero still leads to high unreliability, due to compounding effect of subtle non-determinism over tokens and turns.

In theory, greedy decoding (\textit{i.e.}, $T=0$) will always pick the argmax over the vocabulary distribution.
However, it is reported that hardware limitations on floating point operations cause slightly different intermediate values, which results in a ripple effect of larger value changes and therefore different tokens being selected.

Notable model providers acknowledge the non-determinism implicitly or explicitly; Anthropic recommends sampling multiple times to cross-validate output consistency,\footnote{\url{https://docs.anthropic.com/en/docs/test-and-evaluate/strengthen-guardrails/reduce-hallucinations}.} Google also highlights that their model outputs are \textit{mostly} deterministic,\footnote{\url{https://cloud.google.com/vertex-ai/generative-ai/docs/learn/prompts/adjust-parameter-values\#temperature}.} and OpenAI recommends setting \texttt{seed} parameter to further reduce the non-determinism.\footnote{\url{https://platform.openai.com/docs/advanced-usage\#reproducible-outputs}.}

Nevertheless, we caution users that multi‑turn conversations can be increasingly unreliable owing to divergent LLM responses.

\newpage
\section{Prompts}\label{app:prompts}
\subsection{Sharding}\label{sharding_prompt}
We show the prompts for the sharding process below, using Math as an example task. Double-bracketed terms are placeholders that get replaced with the actual data.
Other tasks share the same outline with different exemplars and rules to enforce stable outputs. We refer the readers to the GitHub repository for the exact prompts on other tasks.
\begin{tcolorbox}[title=Segmentation,float, floatplacement=h]
{\footnotesize
\begin{Verbatim}[breaklines,breaksymbol=]
You are a given a fully specified instruction, and your task is to segment the instruction into a units of information that each reveal a single piece of information of the instruction.
You must output a list of segments in the following JSON format:
[
    {"segment": "[exact excerpt from the instruction]"},
    {"segment": "[exact excerpt from the instruction]"},
    ...
]

Rules:
- [Non-overlapping] The segments must be non-overlapping and cover the entire instruction. You can optionally leave some gaps for non-essential portions of the original instruction (delimiters, headers, etc.)
- [Minimalistic] You should split the information in the segments to as small as possible. If you have a compound expression (X and Y), you should split it into two segments. Each segment should represent a unit of information.

Example Query:
What are the names and locations of the stadiums that had concerts that occurred in both 2014 and 2015?

Output:
{"segments": [
    {"segment": "names and locations"},
    {"segment": "stadiums"},
    {"segment": "concerts"},
    {"segment": "in both 2014"},
    {"segment": "and 2015"}
]}

Now complete the task for the following fully specified instruction:

[[INSTRUCTION]]
\end{Verbatim}
}
\end{tcolorbox}

\newpage
\begin{tcolorbox}[title=Rephrasing]
{\footnotesize
\begin{Verbatim}[breaklines,breaksymbol=]
You are given segments of a fully specified instruction, and your task is to: (1) choose one that will be the initial shard of a multi-step query, and then (2) rephrase each segment into a conversational version that are provided to the system in a follow-up turn of the conversation.

Your output should be a JSON object in the following format:
{
    "initial_segment": "[exact excerpt from the instruction]",
    "initial_shard": "conversational version of the initial segment",
    "shards": [
    {"segment": "[exact excerpt from the instruction]", "shard": "conversational version of the segment taking the rest of the instruction into account"}
    ]
}

Example:

Full Query:
What are the names and locations of the stadiums that had concerts that occurred in both 2014 and 2015?

Segments:
[
    {"segment": "names and locations"},
    {"segment": "stadiums"},
    {"segment": "concerts"},
    {"segment": "in both 2014"},
    {"segment": "and 2015"}
]

Output:
{
    "initial_segment": "stadiums",
    "initial_shard": "popular stadiums",
    "shards": [
        {"segment": "concerts", "shard": "the stadiums should have concerts during a period"},
        {"segment": "in both 2014", "shard": "the concerts should have occurred in 2014 in the stadiums"},
        {"segment": "and 2015", "shard": "the concerts should have also occurred in 2015 in the same stadiums"},
        {"segment": "names and locations", "shard": "for the stadiums, returned both the name and location"}
    ]
}

Rules:
- [Transform each segment] Make sure each segment is included either as the initial shard or in the rest of the shards. Do not forget any segments.
- [Short initial shard] Make the initial shard short, not a full sentence, similar to how users use a search engine like Google.
- [Order of shards] Order the shards in order of importance, from most to least important to the initial shard. You do not need to keep the order the segments that are provided in.

Now complete the task for the following fully specified instruction and segments:

Fully Specified Instruction:
[[QUESTION]]

Segments:
[[SEGMENTS]]

\end{Verbatim}
}
\end{tcolorbox}

\begin{tcolorbox}[title=Verification]
{\footnotesize
\begin{Verbatim}[breaklines,breaksymbol=]
You are given an instruction that fully specifies a problem, and a list of shards. Your task is to decide whether all the information from the full instruction is captured by the shards.

If not, you should output the information unit from the instruction that is not captured by the shards.

Example 1:

Instruction:
What are the names and locations of the stadiums that had concerts that occurred in both 2014 and 2015?

Shards:
{"initial_segment": "stadiums", "initial_shard": "I'm looking for active stadiums", "shards": [{"segment": "concerts", "shard": "the stadiums should have concerts during a period"}, {"segment": "in both 2014 and 2015", "shard": "the concerts should have occurred in both 2014 and 2015"}, {"segment": "names and locations", "shard": "for the stadiums, returned both the name and location"}]}

Output:
{"converage": "complete"}


Example 2:
Instruction:
Which Asian countries have a population that is larger than any country in Africa?

Shards:
{"initial_shard": "I'm interested in learning about countries in Asia", "shards": [{"shard": "consider the population size of these Asian countries"}, {"shard": "the population should be compared in size"}, {"shard": "specifically, compare to the population of African countries"}]}

Output:
{"coverage": "incomplete", "missing_segment": "the shards do not specify that the population of the Asian countries should be *larger* than the population of any African countries"}



You must output in JSON format as shown in the examples above.
Now complete the task for the following fully specified instruction and shards:

Instruction:
[[QUERY]]

Shards:
[[SHARDS]]
\end{Verbatim}
}
\end{tcolorbox}

\newpage
\subsection{Experiments}\label{experiment_prompt}

The experiments involve several LLM calls with specific prompts to simulate the conversation, which we list below.
We refer readers to the GitHub repository for how they are incorporated.

\begin{tcolorbox}[title=User simulator,float,floatplacement=h]
{\footnotesize
\begin{Verbatim}[breaklines,breaksymbol=]
You are simulating a user of an interactive LLM system (like ChatGPT).
The user is inherently lazy, and answers in short form, providing only minimal information to the system. You should not be proactive.

Here's the conversation so far:
[[CONVERSATION_SO_FAR]]

Here are the shards that have already been revealed:
[[SHARDS_REVEALED]]

Here are all the shards that have not been revealed yet:
[[SHARDS_NOT_REVEALED]]

You must generate a response to the conversation so far. Here are the rules:
- [Providing a shard] You can reveal the content of a shard to the system in your response if it will help the system move closer to answering the problem. You should select the shard to reveal that is most "basic" and currently the most relevant.
- [One Shard at a Time] You should only reveal at most one shard at a time.
- [Reveal Entire Shard] If you reveal a shard, you must make sure to include *all the information in the shard*. For example, if the shard is "your symptoms are that you have a headache in the mornings", your response can't just be ``yeah I have headaches'', you must say ``yup mostly headaches in the mornings``.
- [Irrelevant Clarifications] If the system asks you a question irrelevant to the shards, asks you a generic question (``Can you give me a hint?``), you should respond with an answer that does not provide a shard. (``I don't know``, ``Is that really important?``, etc.) You should not reveal any information beyond what is available in the shards.
- [No Repeated Shards] You should not reveal the same shard more than once. Carefully review the already revealed shards, and only reveal a shard if its `shard_id` is not on the list.
- [Rephrase Shards] If you reveal a shard, you should rephrase it in a conversational way. Do not copy the shard verbatim.
- [Do Not Ask Questions] Your response should always be declarative sentences, and not questions.
- [Brevity of Response] You should favor being succint. Your answer can also have typos, improper grammar, capitalization, etc. You are simulating a real person talking to an AI, who is in a hurry.
- [Format] Your response should be formatted as a JSON object with the following keys:
    - `response`: The response to the conversation so far.
    - `shard_id`: The shard you are revealing to the system. The shard_id can be an integer, or -1 if you did not reveal any shards.
For example:
{"response": "I don't know", "shard_id": -1}
or:
{"response": "yeah I want it to [...]", "shard_id": 1}
\end{Verbatim}
}
\end{tcolorbox}
\newpage
\begin{tcolorbox}[title=Response strategy categorization]
{\footnotesize
\begin{Verbatim}[breaklines,breaksymbol=]
You are reviewing a multi-turn conversation between a user and an assistant, and are given the last turn of the conversation.

Here is the full specification of the problem the system is attempting to solve:
[[INITIAL_SHARD]]

Specification:
[[SHARDS]]

You must classify the response of the assistant according to the response type:
- `answer_attempt`: The response contains a complete answer attempt to the user's question (not templated or hypothetical), that can be extracted verbatim. See the task-specific answer description for more details.
- `clarification`: The response  is short (less than 100 words) and contains a single question addressed to the user that directly inquires about an aspect of the user's query. A clarification turn cannot be long (see `discussion`), cannot contain a vague question (see `discussion`) and cannot contain multiple questions (see `interrogation`).
- `interrogation`: The response contains multiple questions addressed to the user, sometimes organized in a list or bullet-points.
- `discussion`: The response discusses the question in detail, without providing a final answer, asking a specific clarification question, or a refusal to answer. The response may or may not contain a vague question (e.g., “What else can I help you with?”).
- `hedge`: The response contains multiple answer candidates based on hypotheticals (ifs) or branching (case 1, case 2) with corresponding descriptions.
- `refuse`: The response contains an explicit or implicit refusal to answer the user's question without a follow-up question or a request.
- `missing`: The response is empty/blank.

You must output your answer in the following JSON format:
{"response_type": "refuse|missing|answer_attempt|hedge|clarification|interrogation|discussion"}

Rules:
- The assistant giving a hint at how an answer could look like is not a final answer. You should only select `answer_attempt` if the conversation could end at this stage with the user having an entirely final answer to the problem they've formulated.
- [Task Specific Answer] [[ANSWER_DESCRIPTION]]

Conversation's last turn:
[[CONVERSATION_SO_FAR]]
\end{Verbatim}
}
\end{tcolorbox}

\begin{tcolorbox}[title=Answer Extraction]
{\footnotesize
\begin{Verbatim}[breaklines,breaksymbol=]
You are reviewing a multi-turn conversation between a user and an assistant, and are given the last turn of the conversation.
In the final response from the assistant, a final answer has been provided. Your goal is to extract verbatim what the answer is:
- If the answer is short (less than 10 words), then you should copy verbatim what the answer is in the `answer` field.
- If the answer is long, then you should produce the answer with an ellipses, to indicate the exact start and end of the answer (e.g, ```def funny_function(n): [...]  return funny_output```). You should include *at least* 4 words or one full line for the start (before the ellipses) and *at least* 4 words or one full line for the end (after the ellipses), such that the answer can be identified exactly.

Rules:
- [Exact Answer Only] only extract the exact answer, and nothing else (including ``` for code blocks, or intro/outro text).
- [Verbatim Only] Only extract verbatim text, do not modify the text in any way. If there's a typo, an error, you must absoltutely include it, and not correct it in any way.
- [Task Specific Answer] [[ANSWER_DESCRIPTION]]
- [String output] the <answer_str> must be a string, not a number and not a dictionary.

You must output your answer in the following JSON format:
{"answer": "<answer_str>"}

Conversation's last turn:
[[CONVERSATION_SO_FAR]]
\end{Verbatim}
}
\end{tcolorbox}

\end{appendices}

\end{document}